\documentclass{article} % For LaTeX2e
\usepackage{iclrlike,times}

\usepackage[utf8]{inputenc} % allow utf-8 input
\usepackage[T1]{fontenc}    % use 8-bit T1 fonts
\usepackage{hyperref}       % hyperlinks
\usepackage{url}            % simple URL typesetting
\usepackage{booktabs}       % professional-quality tables
\usepackage{amsfonts}       % blackboard math symbols
\usepackage{nicefrac}       % compact symbols for 1/2, etc.
\usepackage{microtype}      % microtypography
\usepackage{xcolor}
\usepackage{enumitem}
\usepackage{graphicx}
\usepackage{longtable,multirow}
\usepackage{subfig}
\usepackage{algorithm}
\usepackage[noend]{algpseudocode}
\usepackage{amsmath}
\usepackage{amssymb}
\title{Meta Decision Trees for Explainable\\ Recommendation Systems}

\author{Eyal Shulman \\
Tel Aviv University\\
\And
Lior Wolf \\
Tel Aviv University and\\
Facebook AI Research \\
}

\iclrfinalcopy

\begin{document}

\maketitle

\begin{abstract}
We tackle the problem of building explainable recommendation systems that are based on a per-user decision tree, with decision rules that are based on single attribute values. We build the trees by applying learned regression functions to obtain the decision rules as well as the values at the leaf nodes. The regression functions receive as input the embedding of the user's training set, as well as the embedding of the samples that arrive at the current node. The embedding and the regressors are learned end-to-end with a loss that encourages the decision rules to be sparse. By applying our method, we obtain a collaborative filtering solution that provides a direct explanation to every rating it provides. With regards to accuracy, it is competitive with other algorithms. However, as expected, explainability comes at a cost and the accuracy is typically slightly lower than the state of the art result reported in the literature.
\end{abstract}

\section{Introduction}
There is a growing demand for AI systems that justify their decisions. This is especially the case in recommendation systems (RS), for which users can feel harmed or offended by the provided recommendations. Unfortunately, most RS to date are uninterpretable black boxes. 

Decision trees, with decision rules that are based on single attribute values, are perhaps the most explainable machine learning model in use. The explanation is given by the sequence of decisions along the path in the decision tree taken for a given input sample and each decision in the sequence is directly linked to an input feature. 

\begin{figure}
\centering
\includegraphics[width=0.9996891\linewidth]{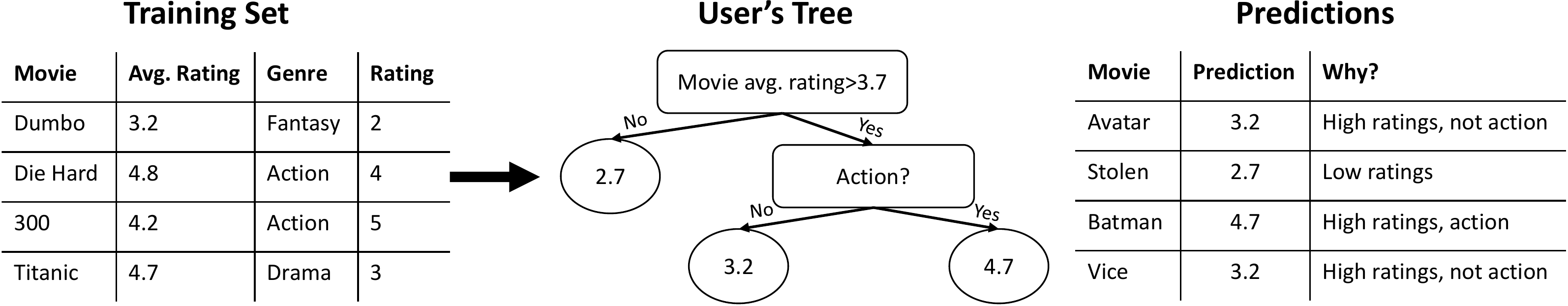}
\caption{A schematic illustration of our explainable recommendation system.}
\label{fig:illustration}
\end{figure}

In the context of RS, fitting a decision tree to predict the user's rating of a given item, based on features derived from the item's and the user's data, leads to uncompetitive performance, due to the task's inherent complexity. Instead, we employ a personalized decision tree for each user, see Fig.~\ref{fig:illustration} for a high-level illustration.  However, learning such a tree based on the limited training data of every single user would lead to overfitting.

Instead, we propose to build a regression function $f$ that maps, at each node, the relevant training samples to a decision rule. Similarly, we train a second network $g$ to learn the value at the leaf nodes. Both $f$ and $g$ are trained end-to-end together with an embedding function $h$ that represents each training sample (both the attribute vector and the provided target value) as a vector. This training includes a data loss on the target values, as well as a sparsity term that encourages the decision rules to be similar to decision stumps. 
The learned functions $f,g,h$ are trained for multiple users in the training set, and play the role of sharing information between users. This is a role that is played in factorization-based collaborative filtering methods by forming basis vectors for the per-user columns of the rating matrix.

Once trained, the decision rules are transformed to decision stumps, which consistently leads to an improvement in performance. Additionally, for maximal explainability, the soft routing that is employed during training is replaced by hard routing, which  somewhat reduces the accuracy. 

We explore two options, using either a fixed or a dynamic architecture. In the fixed architecture, we employ full trees of a certain depth. In the dynamic architecture, we stop the tree growing at empty nodes. On average, the dynamic trees are deeper yet employ fewer decision rules. 

We test our method on all the public recommendation benchmarks we could find. It is demonstrated that our method is competitive with the classical methods of the field. However, the demand for having an explainable decision, especially one that is as straightforward as a decision rule, means that the method falls slightly short in accuracy in comparison to the latest state of the art results.

Our main contributions are: (i) proposing a paradigm for personalized explainable RS, (ii) a learning method for building interpretable recommenders that significantly outperform models with the same level of explainability, (iii) suggesting various model flavors with trade-offs between performance and explainability and (iv) a thorough comparison to multiple algorithms on multiple datasets.

\section{Related Work}

 There is a growing literature on the subject of learning interpretable models,  see \cite{Guidotti:2018:SME:3271482.3236009} for a survey that focuses on generating interpretations for black-box models. 
 Decision trees are one of the few types of models that are naturally explainable, as long as the decision rules are easy to interpret. Our work generates semi-global models, which are fixed per user, in contrast to local explainable models, such as LIME~\citep{Ribeiro:2016:WIT:2939672.2939778} which differ between individual decisions, thus failing to capture the user's model (appendix~\ref{sec:A}).  Local models have also been criticized for not being robust to small modifications of the input ~\citep{yeh2019sensitive, alvarez2018robustness}.
 
Recently,~\cite{alvarez2018} have suggested a framework for explainable neural network models that are monotonic and additive.

The model learns a set of attributes that are trained to be interpretable and
 the explanation takes the form of presenting the contribution of each attribute, while explaining the attributes using prototypes. In our work we rely on decision trees and not on (generalized) linear models.
 We build decision rules with additive contributions from the individual features, but encourage sparseness and advocate for the usage of decision stumps.

In the field of RS, there have been various studies~\citep{tintarev2007survey, zhang2018explainable} showing that equipping recommendations with personalized explanations helps to improve the system's persuasiveness and the users trust in it. In recent years, collaborative filtering methods, in particular those utilizing deep learning models~\citep{hartford2018deep, berg2017graph, zheng2016neural}, have improved recommendation performances, yet, the black box nature of deep learning models and the latent space used in collaborative filtering, make these models hard to explain. 

An explainable RS was presented by~\cite{abdollahi2016explainable,abdollahi2017using} where the recommendations are based on user/item ratings and not on the attributes of the item. The method adds a regularization term to the matrix factorization algorithm, such that users who rated similar items have closer representations. The distribution of each item's rating among the user's nearest neighbors is then provided as an explanations. This is an indirect explanation, which can  be seen as inducing prejudice, by grouping users together.  

\cite{wang2018tem} employed gradient boosted trees as feature extractors, where each feature represents a leaf in a tree. All features, users and items are mapped to a latent space and the final prediction is a linear combination of some bias, a weighted inner product of the user's and item's latent vectors and a weighted sum of an attention-based pooling of the features' latent vectors. Recommendations are explained by the features that have the highest attention scores. These explanations pertain to only a part of the score and are given in the form of a list of complex features, each of which is obtained as a logical formula on top of the input features.

The distillation of the information in a deep network into a soft decision tree~\citep{irsoy2012soft} was studied by~\cite{frosst2017distilling}. Like our work, the decision nodes are characterized by three parameters $w,b,\beta$, and they also observe an improved accuracy with soft assignment to the leaves, which is controlled by $\beta$. In their case, $w$ is not a pseudo probability vector and no sparsification takes process. 
Their decision rules are obtained by minimizing the entire loss, unlike our use of regression functions.

Combining decision trees and neural networks was also studied by~\cite{tanno2018adaptive}, in which trees of dynamic architectures are grown such that the underlying network features gradually evolve.
In our work, the decision rules are based on the original features, and the neural networks we train replace training the trees directly on the labeled set. \cite{tanno2018adaptive} provide an extensive survey of previous work that links decision trees and neural networks. As far as we can ascertain, none of the literature methods employs regression networks for obtaining the decision rules and leaf values.

Our method belongs to the family of meta-learning algorithms, and specifically to the sub-family of few-shot learning~\citep{hinton1987using,schmidhuber1987evolutionary,thrun1998learning,fei2006one,mishra2018a}, in which sample-efficient training for a new task is performed based on observing similar tasks during training. 
The embedding step of our inner loop resembles the embedding step of \cite{mishra2018a}. However, the resulting embedding is used differently in their work, where a learned classifier receives the embedding of the samples as a conditioning input.

\section{Method}

Let $T_h$ be the hypothesis set of all decision trees of depth $h$ with decision rules (inner nodes) of the form $w^\top x \geq b$ for some parameters $w\geq 0$,$b$.  We consider two types of trees, the decision tree itself $t$, and a soft tree, which we denote as $S(t)$, in which each sample is soft assigned by the probability $\sigma(\beta(w^\top x-b))$, for some parameter $\beta$, and these probabilities are multiplied along the path from the root. The leaves contain fixed values, e.g., a score in the case of RS, or some label. In addition, for every tree $t$ we consider a sparsified version $R(t)$, in which the hyperplane type decision rule, based on $w$, is replaced by a decision stump type of rule, in which $w$ is replaced with a one-hot vector that corresponds to the largest value in $w$.

\subsection{The Inner Loop: Tree Generation for a Single User}
\label{sec:inner}

The inner loop generates a tree $t$ given the samples $L$ of a single user $L=\{(x_1, y_1),(x_2, y_2),..,(x_n, y_n)\} \subset \mathbb{R}^{d_x} \times \mathbb{R}^{d_y}$. In other words, it acts as a tree learning algorithm, however, unlike conventional algorithms, it learns how to build suitable trees based on users in the training set, see Sec.~\ref{sec:outer}. Three learned functions are used to build trees: $h: X \times Y \rightarrow \mathbb{R}^{d_h}$, which embeds a training sample and its label, $f$, which returns a decision rule given a set of embedded training samples, and $g$, which provides a leaf value given such a set. 

Within the inner-loop, the information in $L$ is used to build a decision tree $t \in T_h$. This is done using a learned embedding function $h$:
\begin{equation}
r_i = h(x_i, y_i)
\end{equation}

In order to represent the entire set, we employ mean pooling 
$r = \frac{1}{n}\sum_{i=1}^n r_i$.
Given a decision tree $t$, we can grow a new tree node by examining the subset of indices $I\subset[1,\dots,n]$ of samples in $L$, which are assigned to this node. The subset $I$ is also represented by performing mean pooling 
$r_I = \frac{1}{|I|}\sum_{i\in I} r_i$.

The parameters of the decision rule, concatenated to one vector $[w,b,\beta]$ (recall that the third parameter is the softness parameter), are then given by:
\begin{equation}
[w,b,\beta] = f(r,r_I)
\end{equation}
We restrict the vector $w$ to be a vector of pseudo-probabilities, by applying a softmax layer at the relevant head of network $f$. The motivation for this decision is that we view it as a distribution over the various features. In addition, a projection by a vector $w$ that contains multiple signs is hard to interpret. Note that the decision rule can still indicate both ``larger than'' and ``smaller than'' relations, since $\beta$ can be either positive or negative.

The values at leaf nodes are assigned using a learned function as well, which seems to greatly outperform the assignment of the mean value of all samples that arrive at a leaf (denoted by subset $I$):
\begin{equation}
v = g(r,r_I)
\end{equation}

\begin{figure}
\begin{minipage}[c]{0.67\linewidth}
\includegraphics[width=0.96891\textwidth]{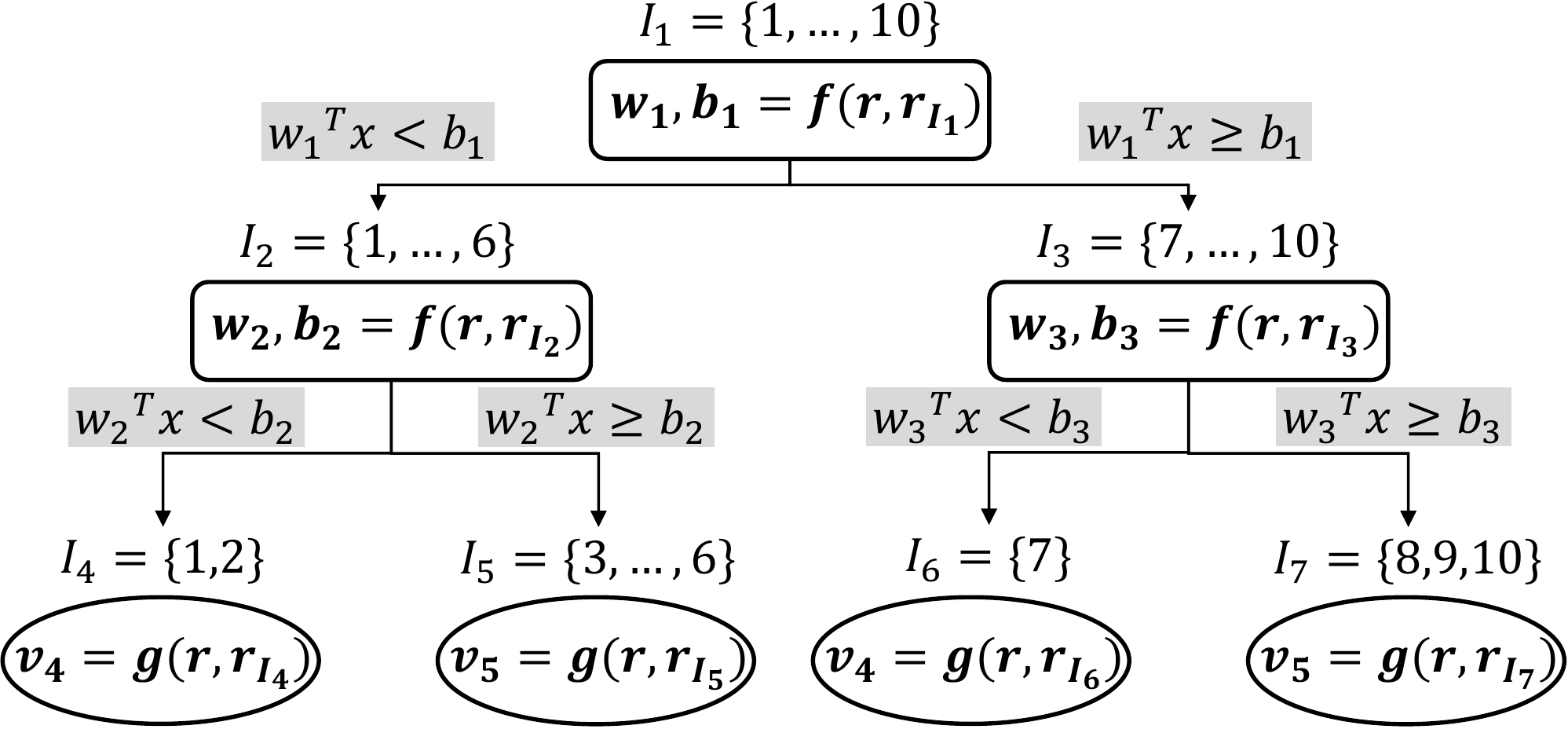}
\end{minipage}%
\hfill
\begin{minipage}[c]{0.32\linewidth}
\caption{An illustration of the method. At the root node, all examples are embedded into $r_{I_1}$, the function $f$ then returns the decision rule to split the samples between the root's right and left children. The process is recursively repeated until reaching leaf nodes, where, using the current set embedding, $g$ returns the leaf value $v$.}
\label{fig:method}
\end{minipage}
\end{figure}

The process of building a tree is illustrated in Fig.~\ref{fig:method} and listed in Alg.~\ref{alg:growtree}. Decision nodes are added in a depth-first manner. Each sample in $L$ is directed to only one node in each tree level, and the time complexity of the method is $\mathcal{O}(nd)$, where $d$ is the maximum depth of the tree. For fixed architecture trees, we employ complete trees of a certain depth. For dynamic trees, nodes in which the decision rule assigns the same label to all training samples that arrive at the node are turned into leaves. 

\noindent{\bf Network architecture} The function $h$ is implemented as a four layer MLP with ReLU activations, in which each layer is of size $d_h$, where $d_h$ is a hyperparameter of the model. The two inputs $(x,y)$ are concatenated at the input layer. Both $f$ and $g$ are implemented as two layer MLPs with a ReLU activation function, with a single hidden layer of size of 20 or 50 respectively. The output of $g$ goes through a logistic (sigmoid) activation function. In the recommendation benchmarks, we linearly scale the output to match the range of target values.

\algnewcommand\Andd{\textbf{and}}
\algnewcommand\Or{\textbf{or}}

\begin{algorithm}[t]
  \caption{The GrowTree procedure. To start the initial tree, we pass $I=[n]$}
  \label{alg:growtree}
  \hspace*{\algorithmicindent} {\textbf{Input} samples $L=\{(x_i,y_i)\}_{i=1}^n$}, boolean $\text{isDynamic}$ tree type, max depth $d$, $I$ relevant indices \\
  \hspace*{\algorithmicindent} {\textbf{Output} a decision tree $t$}
  \begin{algorithmic}[1]
  \vspace{.1cm}
  \State $r \gets \frac{1}{n}\sum_{i=1}^n h(x_i, y_i)$ \Comment{$r$ and $h(x_i,y_i)$ can be computed once}
  \State $r_I \gets \frac{1}{|I|}\sum_{i \in I} h(x_i, y_i)$
  \If{$d=0$} \Comment{Check if building a leaf node or an inner node}
    \State \textbf{return} $Leaf(g(r, r_I))$\Comment{The value of the leaf node}
  \EndIf
  \State $w, b, \beta \gets f(r, r_I)$ \Comment{The parameters of the decision node}
  \State $I_l \gets \{ i \in I| \beta w^\top x_i < \beta b \}$
  \State $I_r \gets \{ i \in I| \beta w^\top x_i \geq \beta b \}$
  \If{$\text{isDynamic}~~~ \Andd~~~ (I_l=\varnothing~~~ \Or ~~~I_r=\varnothing$)} \Comment{For dynamic trees, stop at empty sets}
    \State \textbf{return} $Leaf(g(r,r_I))$ \Comment{The value of the leaf node}
  \EndIf
  \State $child_l \gets GrowTree(L, isDynamic, d-1, I_l)$
  \State $child_r \gets GrowTree(L, isDynamic, d-1, I_r)$
  \State \textbf{return} $Node(w, b, \beta, child_l, child_r)$
  \vspace{.1cm}
  \end{algorithmic}
\end{algorithm}

\subsection{The Outer Loop: Training the Networks on the Set of Training Users}
\label{sec:outer}

For training the three networks, we obtain a set of labeled training sets $K=\{L_i\}_{i\in[k]}$.  Each set is split into two: $L'_k$ that is used to build a decision tree $t_k$, and $L''_k$ that is used to evaluate the tree and provide an error signal for training the three networks $h,f,g$. Specifically, during training, the trees are constructed using the current version of the networks $h,f,g$ using Alg.~\ref{alg:growtree}: 
\begin{equation}
t_k = GrowTree(L'_k,h,f,g).
\label{eq:tk}
\end{equation}

The decision tree $t_k$ is a function from a sample $x$ to some leaf label $v$. For the purpose of computing a loss, we consider the soft decision tree $t^\text{soft}_k = S(t_k)$, in which each sample $x$ is soft-assigned to the leaves of the tree $t_k$. Specifically, if a path to leaf node of depth $\delta+1$ is given by a set of decision rules with parameters ${(w_i,b_i,\beta_i)}_{i=1}^\delta$, the sample is assigned to the leaf with the probability $\Pi_i \sigma(\beta_i(w_i^\top x-b_i))$. The parameter $\beta_i$ controls the softness of the obtained probability.

Let $Z$ be the set of tree leaves, where each leaf $z \in Z$ being assigned a value $v_z$. Let $t^\text{soft}_k(x,z)$ be the soft assignment of $x$ into leaf $z$. The regression loss that is used during training is given by:
\begin{equation}
\label{eq:loss}
    \mathcal{C}(K,h,f,g) = \sum_{(L'_k,L''_k)\in K} \sum_{(x,y)\in L''_k} ||y- \sum_{z\in Z} t^\text{soft}_k(x,z)v_z||^2 
\end{equation}
where the link between the loss and the learned networks is given by Eq.~\ref{eq:tk}. In addition, we would like to encourage sparsity of each decision node in the tree. Let $W(t)$ be the set of hyperplanes $w$ gathered from all nodes of a decision tree $t$ with linear decision rules, and $D$ be the operator that returns the diagonal part of a matrix, we define the following loss:
\begin{equation}
\label{eq:rt}
\mathcal{R}(t) = \frac{1}{|W(t)|}\sum_{w\in W(t)} | ww^\top - D(ww^\top) |_1
\end{equation}
The overall loss is given by 
$\mathcal{L}(K,h,f,g) = \mathcal{C}(K,h,f,g) + \lambda \sum_{k=1}^K  \mathcal{R}(t_k)$.
The first part of the loss is the regression loss, and the second promotes the sparsity of the solution, which pushes the hyperplanes $w$ toward the form of one-hot vectors.

\subsection{Inference: Building a Tree at Test Time}
\label{sec:R}
We obtain the tree $t$ given a user training set $L$ as described in Sec.~\ref{sec:inner}. Since the nodes perform a linear combination of the inputs, they are only partly explainable. For this reason, we replace the tree $t$ with a sparsified tree $R(t)$, in which each node is a decision stump: for a node with parameters $w,b,\beta$, representing the decision rule $\beta (w^\top x - b) \geq 0$, we identify the index $i$ that has the maximal value in $w$, i.e., $i=\arg\max w[i]$ (recall that $w$ is a positive vector). We then replace the decision rule of the node with the decision rule $x[i]\geq \frac{b}{w[i]}$ if $\beta$ is positive, and  $x[i] < \frac{b}{w[i]}$ otherwise.

\section{Experiments}

\noindent{\bf A Synthetic Classification Problem~~} 
We test the meta trees method on a synthetic binary classification problem. The ground truth labels are obtained from a decision tree of depth two, where each node is a decision stump. Each set of samples $L$ in the experiment consists of random $d$-dimensional vectors, and the set's ground truth decision tree is constructed by drawing three random indices $\{i_1, i_2, i_3 \} \subset [d]$ from a skewed distribution defined as $p(i)=\frac{s^i}{\sum_{i=1}^{d} s^i}$, where $s$ controls the skewness of the distribution. Finally, the set's labels are determined as $y=((x_{i_1}>0) \land (x_{i_2}>0)) \lor ((x_{i_1} \leq 0) \land (x_{i_3}>0))$. A separate set unseen during training, classified by the same decision rule, is used for evaluation. In our experiments, $d=10$ and $s=1.3$. In the training set, the number of samples in each set $L$ is sampled uniformly $U(1,50)$.

We test three types of meta trees: (i) full trees of depth two, (ii) full trees of depth three,  and (iii) trees that are grown according to the stopping criterion, up to maximum depth of five. In order to train the meta trees for this classification problem, we replace the squared loss in Eq.~\ref{eq:loss} with the logistic loss. For all three models $d_h=512$ (which seems optimal for the three models) and the loss is minimized by the Adam optimizer with a learning rate of $3 \cdot 10^{-4}$ and batch size of $256$.

We compare sparse meta trees with hard decisions (each sample is associated with a single leaf) to two types of vanilla decision trees, of depths two or three: (i) Local Trees -- a decision tree classifier fitted on the current set $L$ of training samples, and (ii) Global Tree -- a decision tree classifier fitted on the union of the training samples from all training sets, with the addition of the set $L$. Both trees employ decision stumps and are trained by the CART algorithm by~\cite{cart} implementation in scikit-learn~\citep{scikit-learn}.

\begin{figure}[t]%
\centering
\begin{tabular}{ccc}
\includegraphics[width=0.31\textwidth]{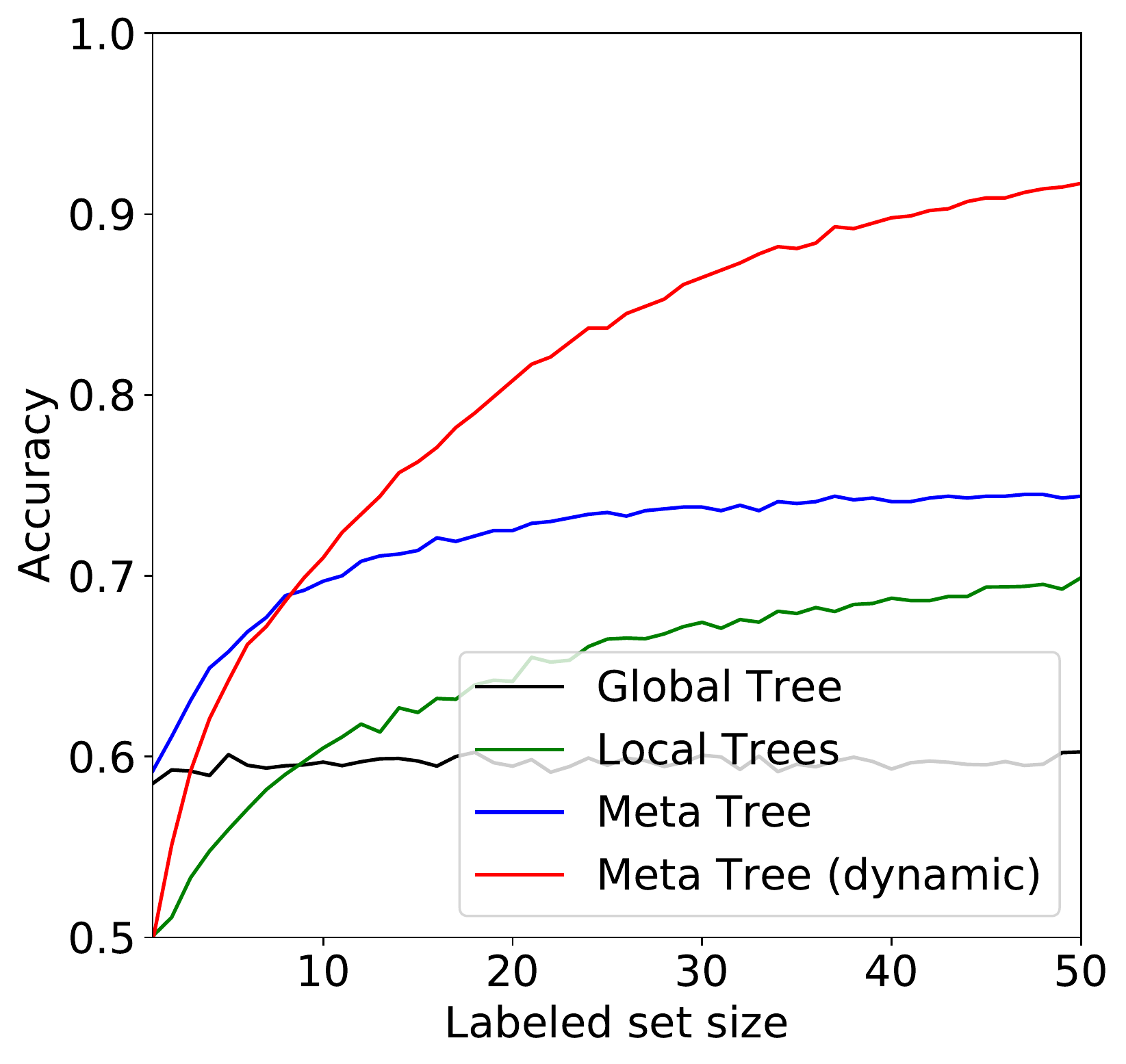}&
\includegraphics[width=0.31\textwidth]{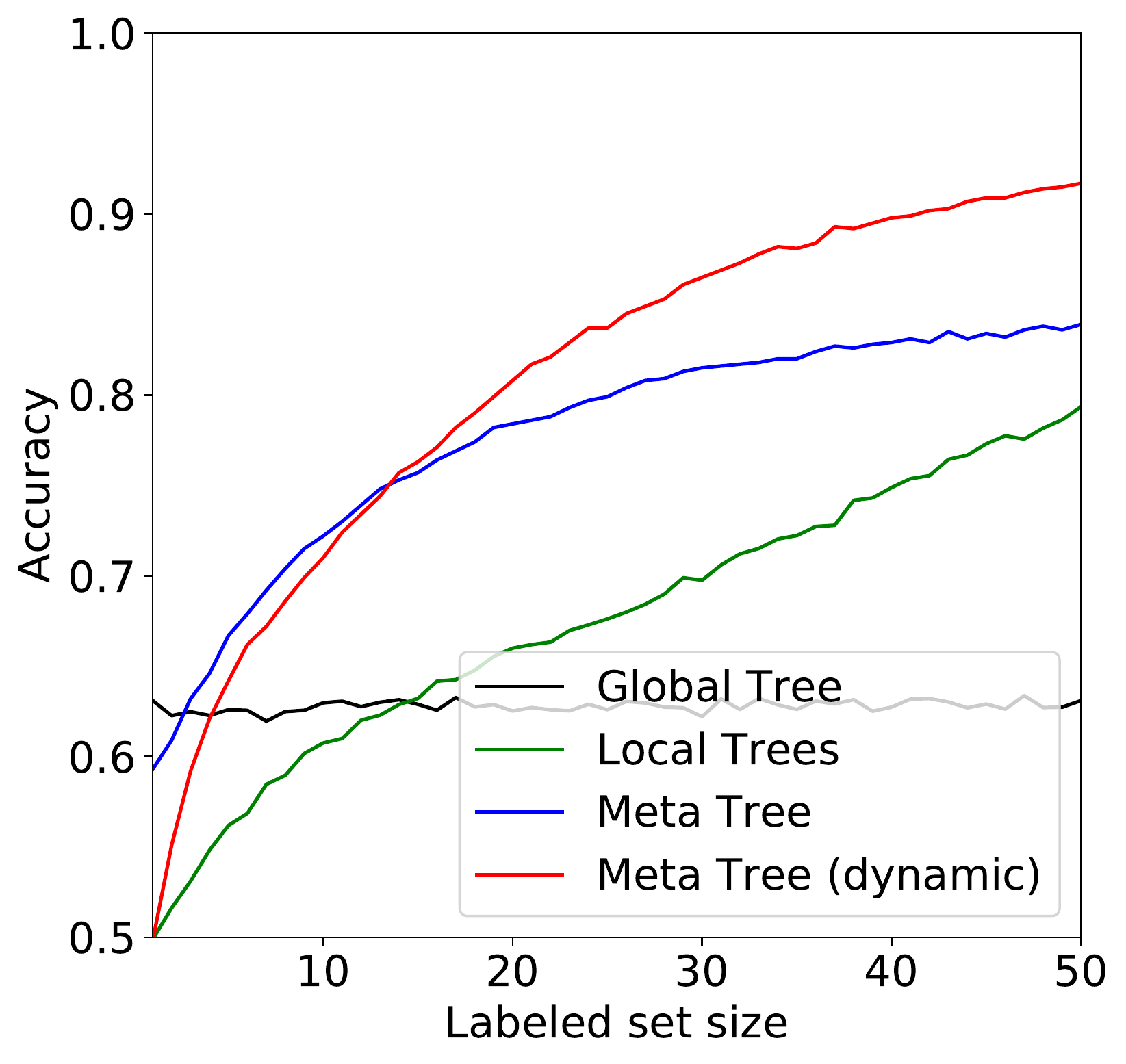}&
\includegraphics[width=0.298\textwidth]{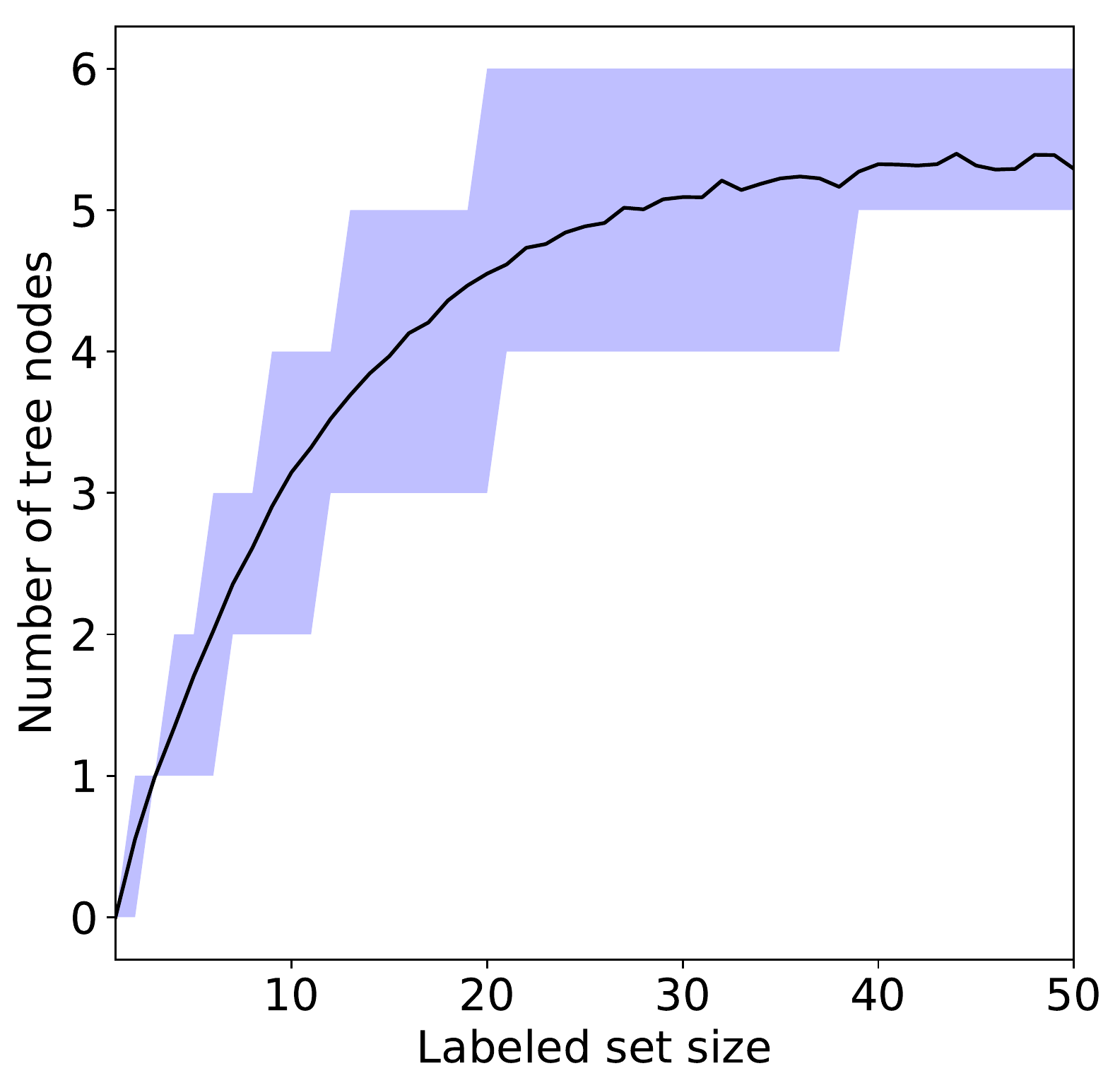}\\
(a)&(b)&(c)\\
\end{tabular}
\vspace{-.34cm}
\caption{Synthetic problem. (a,b) Performance as a function of the labeled set size ($|L'|$). The depth of all models, except for the dynamic meta tree, is 2 in panel (a) and 3 in panel (b). (c) The average number of inner nodes, as well as the 25th and 75th percentiles, used by the dynamic trees.}%
\label{fig:synthetic}%
\vspace{.4cm}
\centering
\begin{tabular}{lcc}
&MovieLens-100k       &   Jester       \\
{{\rotatebox[origin=l]{90}{~~~~~~~~~~~~Global trees}}}&\includegraphics[width=0.4635001\textwidth]{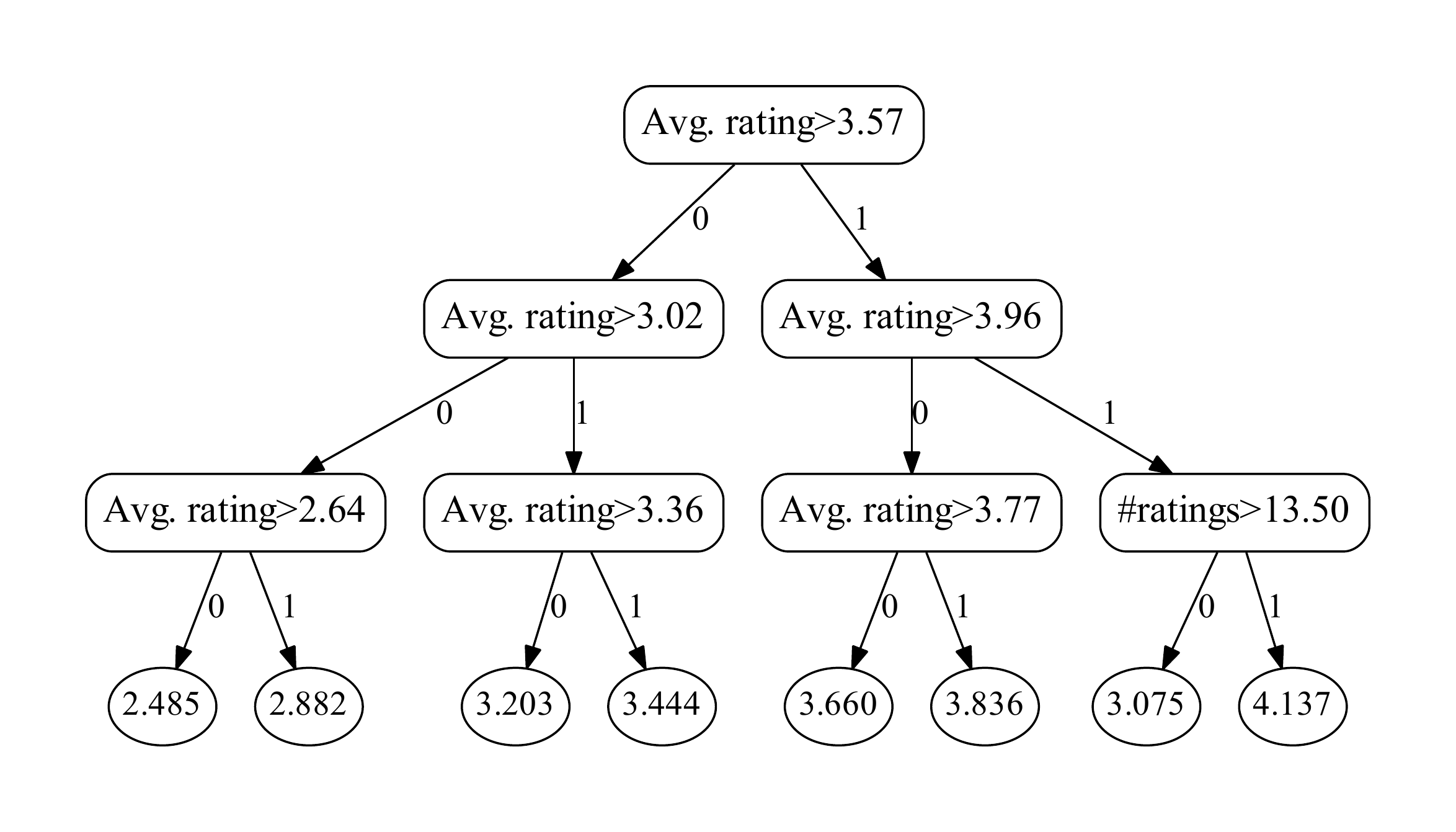}&
\includegraphics[width=0.4635001\textwidth]{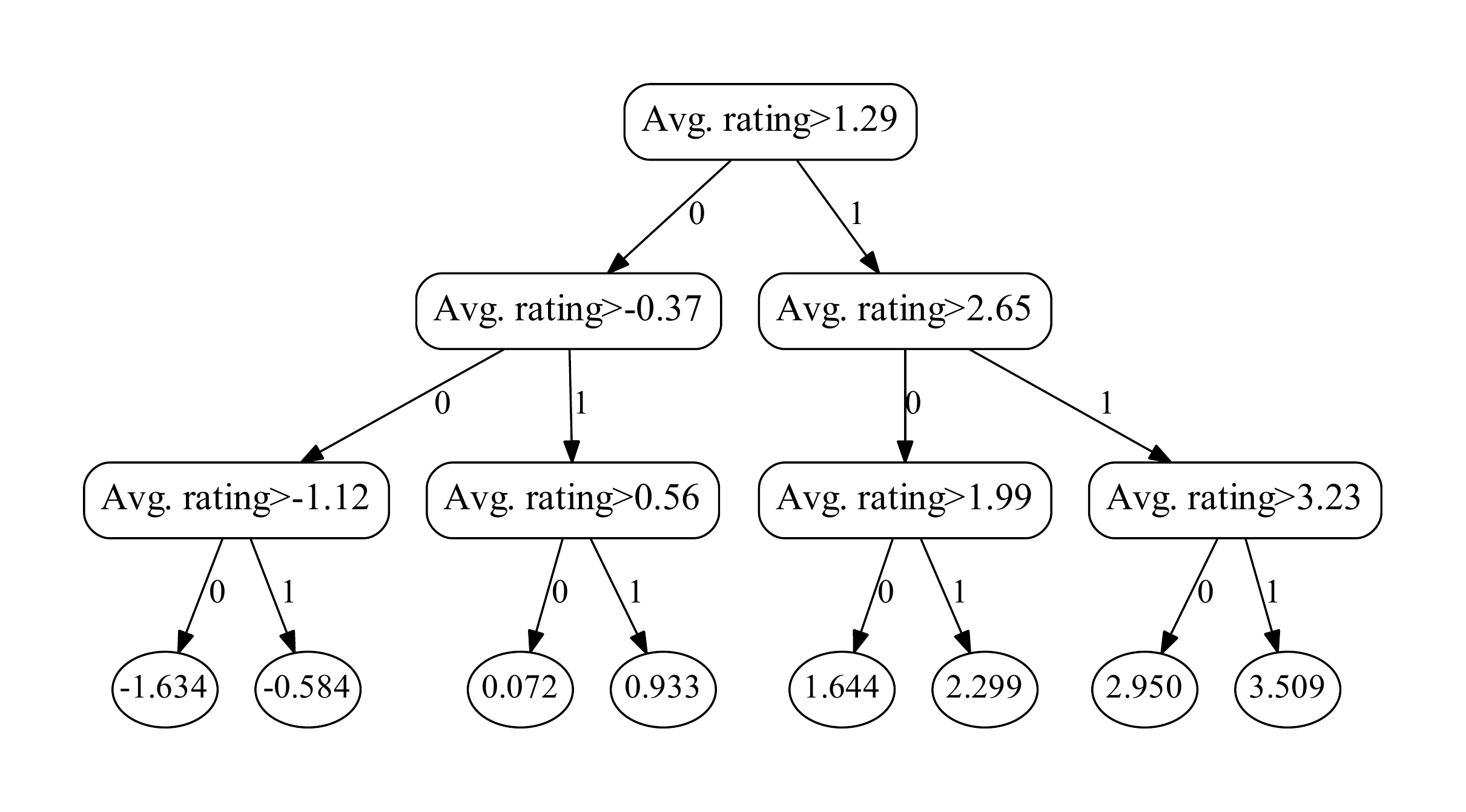}\vspace{-.3cm}         \\
&(a)&(b)\\
{{\rotatebox[origin=l]{90}{Fixed depth meta-trees}}}&\includegraphics[width=0.4635001\textwidth]{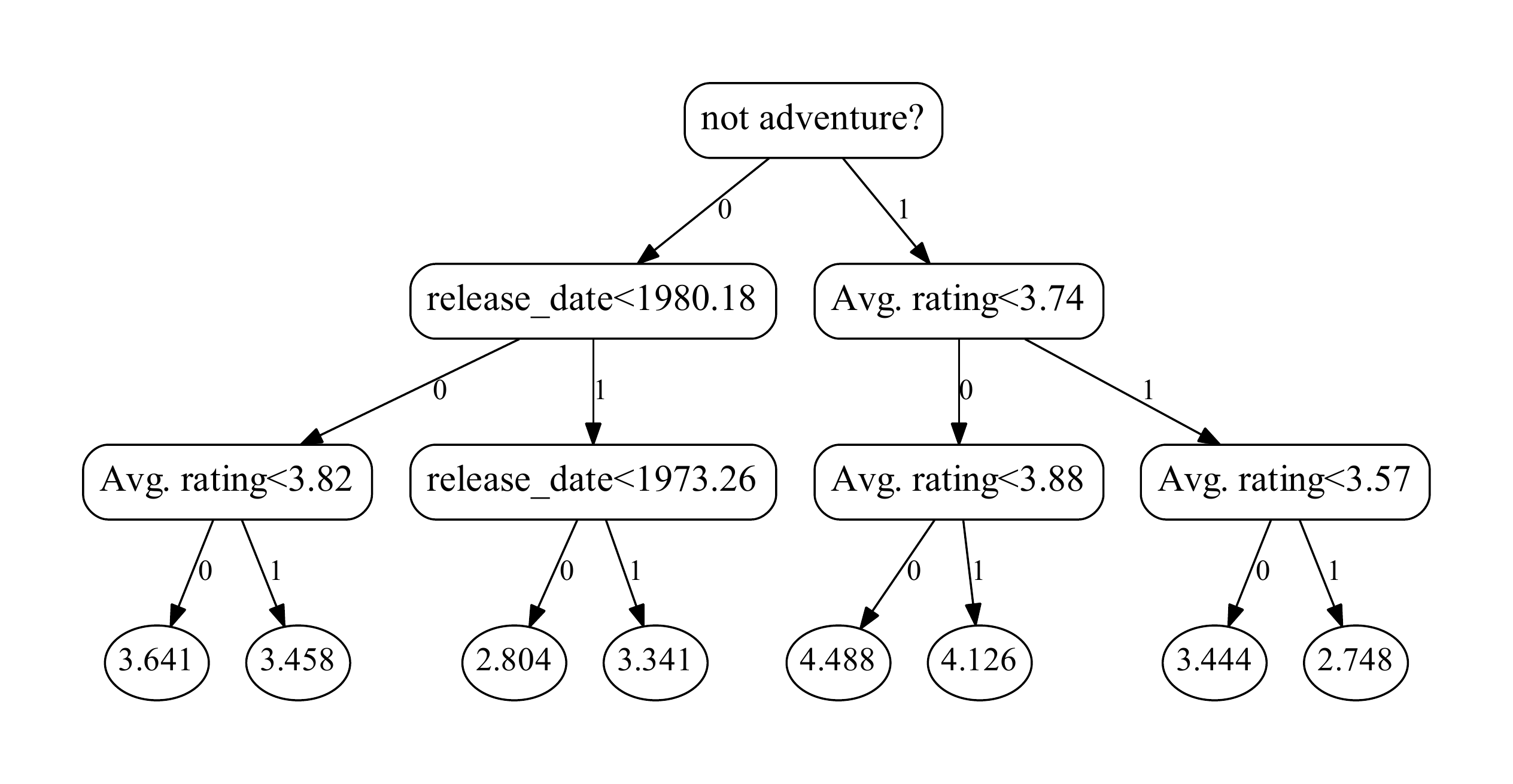}&
\includegraphics[width=0.4635001\textwidth]{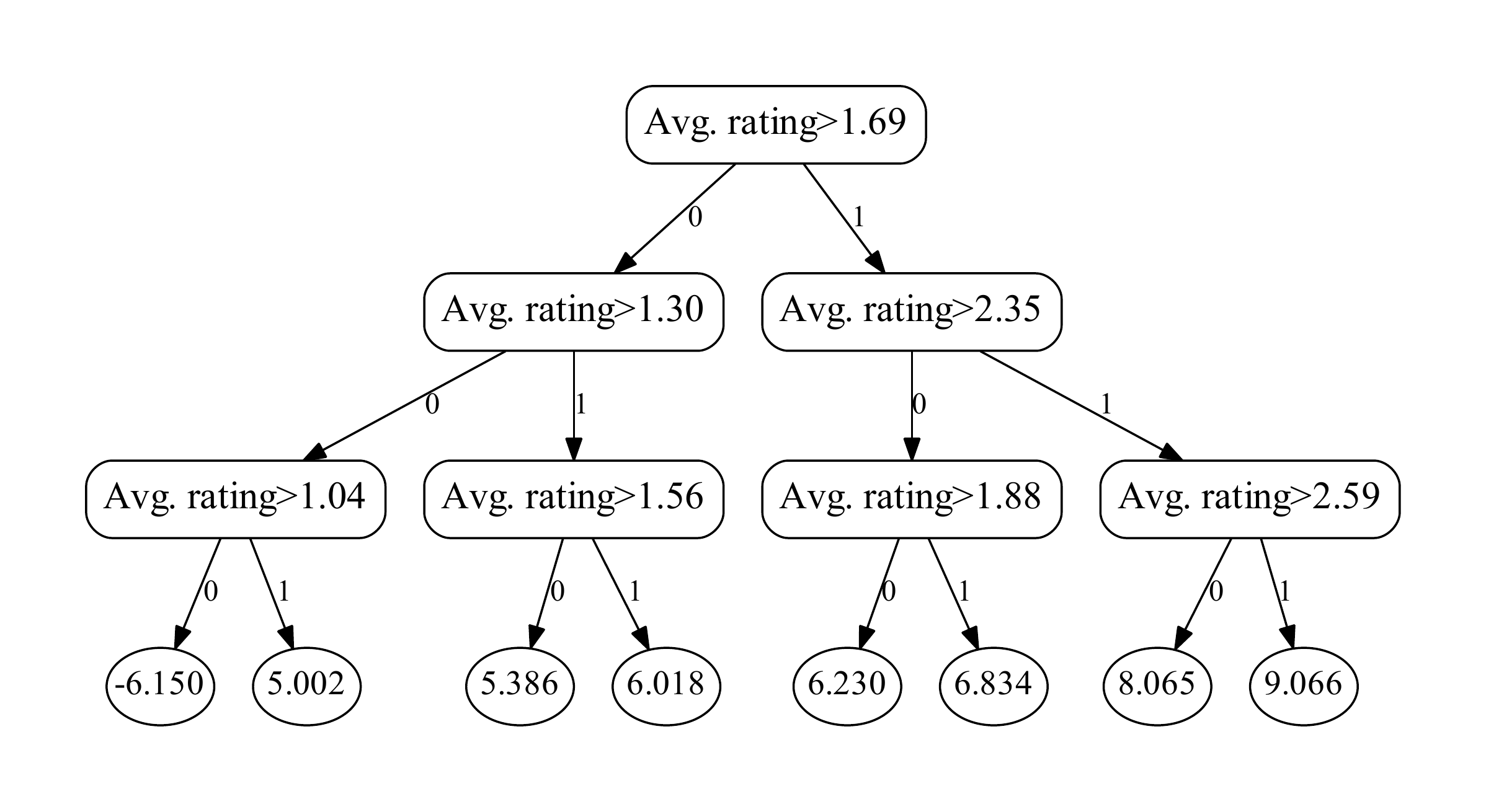} \vspace{-.3cm}        \\
&(c)&(d)\\
{{\rotatebox[origin=l]{90}{Dynamic meta-trees}}} &\includegraphics[width=0.4635001\textwidth,height=0.15\textheight,keepaspectratio]{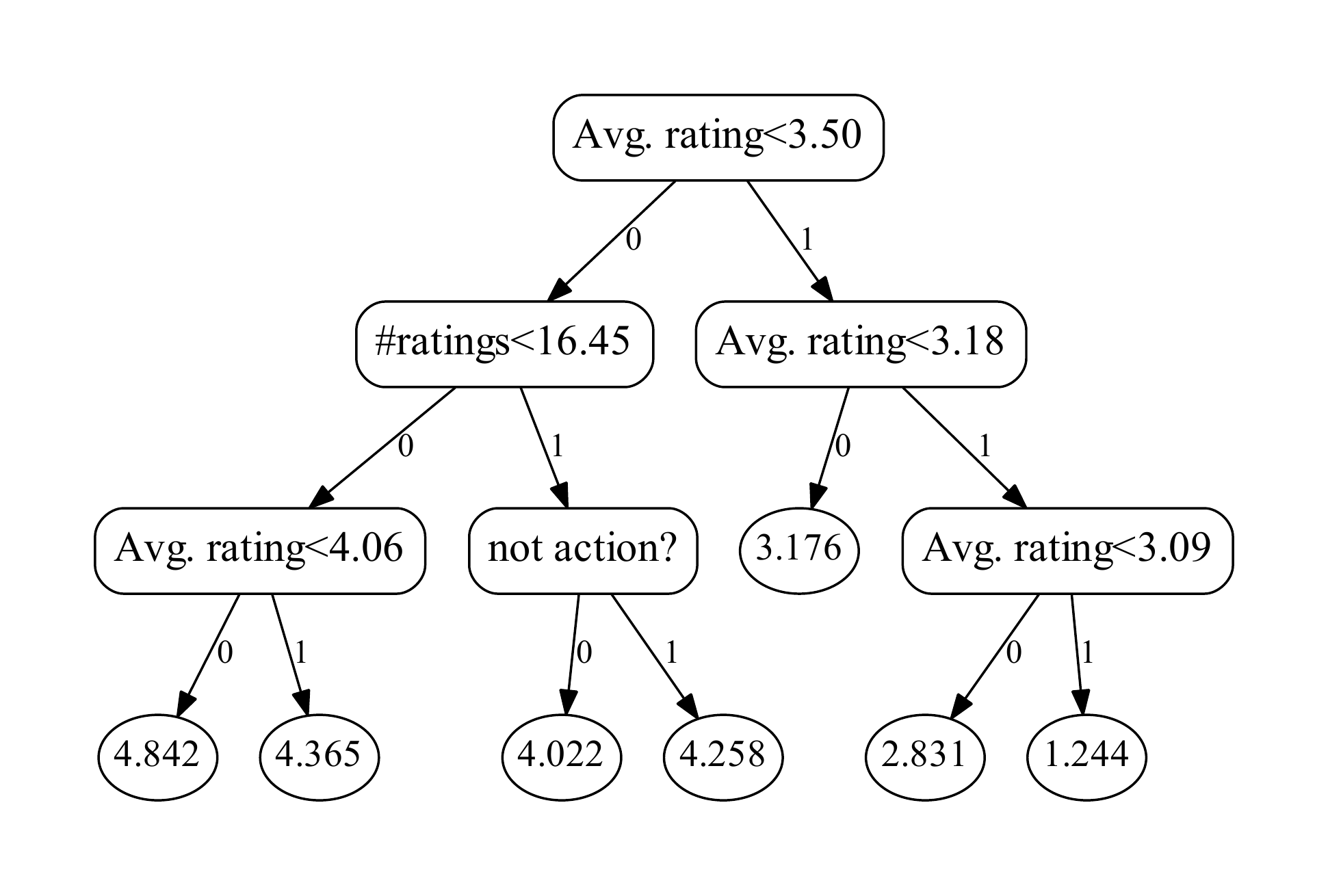}&
\includegraphics[width=0.4635001\textwidth,height=0.15\textheight,keepaspectratio]{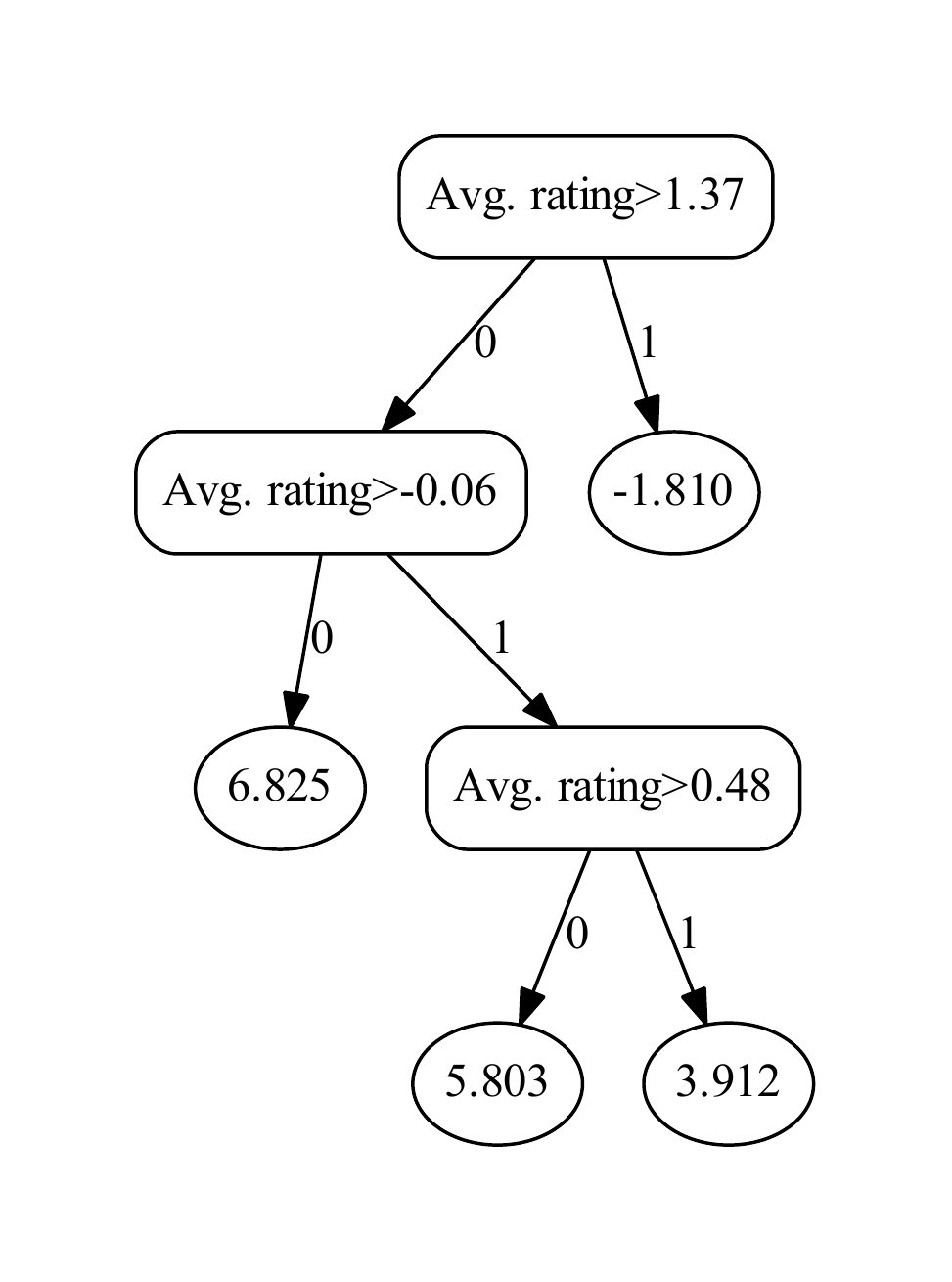}        \vspace{-.3cm} \\
&(e)&(f)\\
\end{tabular}
\caption{Samples of global trees and meta-trees that are either dynamic or of a fixed architecture.}%
\label{fig:trees}%
\end{figure}

The results of the experiment are shown in Fig.~\ref{fig:synthetic}(a,b) for trees of depth two and three, respectively. Note that the dynamic meta-tree is the same experiment in both panels. The results show that the vanilla trees are outperformed by the fixed-depth meta-tree as well as by the dynamic architecture meta tree. As expected, the accuracy of all meta trees models, as well as that of the local trees, increases as more training samples are presented. The results also indicate that the dynamic tree outperforms the fixed tree of depths two or three when more training samples are present. Additionally, while the fixed trees of depth three use exactly seven inner nodes, the dynamic trees tend to employ less. The average number of nodes, as well as the 25th and 75th percentiles, are shown in Fig.~\ref{fig:synthetic}(c).

\noindent{\bf Recommendations datasets~~} To test our method on real data, we use three popular recommendations datasets, MovieLens 100K, MovieLens 1M~\citep{harper2016movielens} and Jester~\citep{goldberg2001eigentaste}. Two other datasets that are sometimes used in the literature, Netflix and Duban, are currently not available for research. The MovieLens datasets consist of integer movie ratings from 1 to 5 and include the movie metadata, such as the movie genres and its release year. Our decision trees are based on this meta data with the addition of the average rating for each movie and the number of times it was rated, both computed on the entire training set. The MovieLens 100K dataset contains 100,000 ratings of 943 users for 1680 different movies, and the MovieLens 1M dataset contains 1 million ratings of 6,040 users for 3,706 different movies. The Jester dataset is of continuous jokes ratings from -10 to 10, containing the jokes' texts. We extracted several features from the jokes texts, such as the number of lines in the joke, the number of all tokens, and the number of occurrences of each of the 50 most common non-stop tokens in the dataset. The dataset contains users with very few ratings, after removing all users with less than 20 ratings, it contains approximately 1.4 million ratings of 26,151 different users for 140 different jokes.

For the MovieLens 100k dataset, we report performance on the canonical u1.base/u1.test. For the MovieLens-1M and the Jester datasets, 10\% of the data was randomly selected for the test set.

\begin{table}[t]
\caption{Performance comparison on the MovieLens 100k, MovieLens 1M and Jester datasets.}
\label{tab:cf_compare}
\vspace{-2mm}
\centering
\begin{tabular}{lcccccc}
\toprule
 & \multicolumn{2}{c}{MovieLens 100K} & \multicolumn{2}{c}{MovieLens 1M} & \multicolumn{2}{c}{Jester} \\ 
 \cmidrule(lr){2-3}
 \cmidrule(lr){4-5}
 \cmidrule(lr){6-7}
                                        & RMSE      & MAE       & RMSE     & MAE        & RMSE      & MAE       \\
\midrule
Global Tree (best depth)                    & 0.995     & 0.778     & 0.935     & 0.738     & 4.302     & 3.136     \\ 
Local Trees (best depth)                    & 1.018     & 0.791     & 0.947     & 0.737     & 4.556     & 3.137     \\ 
KNN Trees (best k)                          & 0.975     & 0.770     & 0.921     & 0.726     & 4.161     & 3.070     \\ 
Gradient boosted regression trees           & 0.976     & 0.771     & 0.933     & 0.736     & 4.119     & 3.131     \\ 
SVD                                         & 0.953     & 0.751     & 0.868     & 0.682     & 4.073     & 2.987     \\ 
SVD++                                       & 0.932     & 0.730     & 0.861     & 0.671     & 4.198     & 3.146     \\ 
GRALS~\citep{rao2015collaborative}          & 0.945     & -         & -         & -         & -         & -         \\ 
sRGCNN~\citep{monti2017geometric}           & 0.929     & -         & -         & -         & -         & -         \\ 
{Factorized EAE}~\citep{hartford2018deep}   & 0.920     & -         & 0.860     & -         & -         & -         \\
GC-MC~\citep{berg2017graph}                 & 0.905     & -         & 0.832     & -         & -         & -         \\ \midrule
Meta Trees (sparse, soft)                   & 0.947     & 0.747     & 0.876     & 0.687     & 4.001     & 3.012     \\
Meta Trees (sparse, hard)                   & 0.970     & 0.766     & 0.916     & 0.722     & 4.131     & 3.030     \\
Dynamic Meta Trees (sparse, soft)           & 0.948     & 0.747     & 0.872     & 0.683     & 4.008     & 3.062     \\
Dynamic Meta Trees (sparse, hard)           & 0.975     & 0.767     & 0.914     & 0.720     & 4.171     & 3.097     \\
\bottomrule
\end{tabular}
\end{table}

We compare the performance of our method to acceptable literature baselines, which include: SVD and SVD++~\citep{koren2008factorization} which are popular collaborative filtering methods implemented in the Python Scikit Surprise package~\citep{hug2017surprise}, various tree baselines, and the state of the art methods for both MovieLens benchmarks. From these methods, only the tree results are explainable. 

The tree baselines include the Global- and Local-Tree models described in the synthetic problem section. 
When presenting the results of these global and local trees, we choose the hyperparameters that demonstrate the best test performance. This is done in order to set an upper bound on their performance. For global (local) trees, we used a maximum tree depth of three, four and three (one, two and one) for the MovieLens 100K, MovieLens 1M and Jester respectively. 
An additional baseline, called kNN trees is added, in which the training set is augmented by the training set of the $k$ nearest users. User similarity is determined according to the cosine distance of the user embedding vectors generated by the SVD++ method. This way, we augment the data of the user, which is limited by nature, by the data of similar users, mitigating the risk of overfitting and potentially obtaining a more accurate tree. The performance shown is for the best possible $k$ as evaluted on the test set. See appendix~\ref{sec:knntree} for a graph depicting the performance as a function of $k$. While less interpretable than the 
other tree-based baselines, we also add the performance of gradient boosted regression trees ~\citep{friedman2001greedy}, using the scikit-learn implementation. We have also implemented the TEM method of~\cite{wang2018tem}. However, despite considerable effort, it failed to produce reasonable results and we believe that the method is especially suited for the case of a large number of attributes.

For all benchmarks, we used a tree depth of three for the fixed architecture meta trees, and for both the dynamic and fixed architectures we used $d_h=512$ and $\lambda=0.1$. This set of parameters was set early on in the development process and kept for all future experiments. In appendix~\ref{sec:paramsensitivity}, we study the parameter sensitivity showing that the model is largely insensitive to its parameters.

The models are compared in terms of the acceptable evaluation metrics: root-mean-square-error (RMSE) and mean-absolute-error (MAE). The results of the experiment are shown in Tab.~\ref{tab:cf_compare}. From the results it can be seen that for both MovieLens datasets the model outperforms the trees benchmarks and achieves comparable performances to those of the SVD algorithm, but often falls slightly behind the latest unexplainable methods. On the Jester dataset, for which most of the recent literature methods do not report results, the model outperforms all baseline methods. See appendix~\ref{sec:coldstart} for an analysis of the model performance and behavior with respect to the users' training set size. In addition, our model is robust to small modifications of the training set, see appendix~\ref{sec:robust}.

White-box models support introspection. 
In all three datasets we found the following types of trees: 
    (i) Feature specific trees: For most users, the model builds trees which use specific features. The users depicted in  
    Fig.~\ref{fig:trees}{\color{black}(c,e)} 
    are generally aligned with the average rating. However, the first dislikes adventure movies and the second tends to trust the average rating when the movies were frequently rated.

    (ii) Classic CF trees: This type of tree splits the examples by their average rating and adjusts the leaf values to the specific user ratings. The model produces such trees mostly for users with a small number of ratings, see appendix~\ref{sec:coldstart}. This behavior is very similar to that of a baseline CF algorithm where the predicted user rating is the sum of the dataset average rating, the item bias and the user bias. An example for such a tree built for a Jester user with very high ratings is shown in Fig.~\ref{fig:trees}(d).

    (iii) Reverse trees: this type of tree is built for users which prefer ``bad'' movies/jokes, i.e. they rate highly items with low average ratings. The trees the model builds for such users are very similar to classic CF trees in terms of splitting the examples. The difference is that the leaf values appear in reverse order, adjusted to the specific user ratings. Fig.~\ref{fig:trees}{\color{black}(f)} shows an example for such a tree. We show a more thorough analysis of this behavior in appendix~\ref{sec:contrarian}.

\noindent{\bf Ablation analysis} To evaluate the various contributions, we have evaluated the following variants of our method: 1. Test on trees without applying the sparsification  operator $R$ first (see Sec.~\ref{sec:R}). 2. Removing the sparsification term (Eq.~\ref{eq:rt}) from the training loss. 3. Using mean value in the leaves instead of $g$. 4. Removing the first input $r$ from $f$ and $g$. 5. Training with a fixed $\beta=1$. 6. Setting $\beta=1$ but allowing $w$ to get negative values. 7. One-hot $w$ in training, using  straight-through (ST) estimators~\citep{st}. 8. Employing hard routing and ST estimators when computing loss. 9. Measuring the loss also on the train samples in $L_k$, following~\cite{mishra2018a}. As Tab.~\ref{tab:ablation} shows, all of these modifications seem detrimental to the loss. For example, when working with trees where the sparsification operator $R$ was not applied, the results for hard routing deteriorate slightly. Using a fixed beta significantly harms the hard routing performance, and removing $g$ results in poor performances for both soft and hard. The only exception is variant 8, which could help improve results for hard routing at inference.

\begin{table}
\caption{An ablation analysis testing various variants of our method (RMSE).}
\label{tab:ablation}
\vspace{-2mm}
\centering
\begin{tabular}{@{}l@{~}c@{~}c@{~}c@{~}c@{~~~}l@{~}c@{~}c@{~}c@{~}c@{}}
\toprule
 &  \multicolumn{2}{c}{M.Lens100} & \multicolumn{2}{c}{Jester}  & & \multicolumn{2}{c}{M.Lens100} & \multicolumn{2}{c}{Jester} \\ 
 \cmidrule(lr){2-3}
 \cmidrule(lr){4-5}
  \cmidrule(lr){7-8}
   \cmidrule(lr){9-10}
Method/Routing                                 & Soft      & Hard      & Soft      &  Hard & Method/Routing & Soft      & Hard      & Soft      & Hard   \\
\cmidrule(lr){1-5}
\cmidrule(lr){6-10}
Meta Trees (sparse)                     & 0.947     & 0.970     & 4.001     & 4.131   &  5. Without $\beta$                      & 0.950     & 1.317     & 4.022     & 5.118     \\
1. Semi-sparse             & 0.948     & 0.974     & 4.035     & 4.131  &   6. W/o $\beta$,allowing $w\!<\!0$   & 0.953     & 0.999     & 4.064     & 5.946     \\
2. No sparse norm                       & 0.953     & 0.988     & 4.024     & 4.190     & 7. 1-hot weights in train             & 0.958     & 1.024     & 4.035     & 4.277     \\
3. Mean leafs(no $g$)            & 1.019     & 1.063     & 4.209     & 4.372     & 8. Hard routing in loss                 & 0.969     & 0.973     & 4.050     & 4.077     \\
4. $g$,$f$ only using $r_I$             & 0.952     & 0.984     & 4.083     & 4.243   &  9. Using train for loss                 & 0.951     & 0.982     & 4.056     & 4.200     \\
\bottomrule
\end{tabular}
\end{table}

\section{Discussion}

Recommendation systems naturally call for meta-learning and inter-user knowledge sharing, and due to the direct implications of the recommendations on users, transparency and interpretability are of high interest. 
This is also true for personalized medicine and credit decisions. However, we could not identify suitable datasets in these domains.

We believe that semi-global explainable models, in which the user (i) has full understanding of her personalized decision rules, (ii) can validate that the model is unbiased toward protected groups, and (iii) can even edit the rules in an intuitive manner, is the right trade-off between explainability and performance. While the tree generation networks behind the scenes are unexplainable, this is unavoidable, since any global explainable model would be inherently inaccurate due to the need to capture a diverse set of users with a model of limited capacity.

\section*{Acknowledgments}
This project has received funding from the European Research Council (ERC) under the European Union's Horizon 2020 research and innovation programme (grant ERC CoG 725974).

\bibliography{cf}
\bibliographystyle{iclr2020_conference}

\appendix

\section{Local, semi-global, and global explainable models}
\label{sec:A}

We extend upon definitions of interpretable notions, which are only briefly discussed in the related work and conclusions sections. A globally interpretable model is one in which the entire logic and reasoning of the model are clear to the user. An example would be a global decision tree, in which the user can understand the logical flow that led to a certain decision. The limiting factor in such models is that the prediction capability is limited by the capacity of the model, which itself is limited by the need to remain interpretable to human users. 

In contrast, local interpretability is the ability to understand the local decision. Local models such as LIME~\citep{Ribeiro:2016:WIT:2939672.2939778} and SHAP~\citep{lundberg2017unified} aim to provide this ability. For every specific decision, the influence of local perturbations on the final decision is analyzed. In the case of a movie rating, the user is able to ascertain from the explanation that for a specific movie,  for example, an increase in the average rating in the population would strengthen the recommendation and that if the movie was not a drama, it would not have been recommended. However, the user cannot tell if such considerations would apply to a different movie (the explanation is valid for only one movie). The user can attempt to obtain a general understanding by examining many samples, but this process is based on the user's predictive capabilities, is subjective, and more importantly, is not given by the model.

In the semi-global model that we present, the global model is divided into explainable parts. The user is able to grasp the model that determines the recommendations for that user, while not being able to understand how the algorithm obtained this model. Such a model is transparent to the user in the sense that the user has a complete understanding how the predicted rating of a specific item is determined and can evaluate the suitability of the model for her personal preferences. Such a model disentangles the user-specific information from that of the entire population and, therefore, does not suffer from the capacity problem as much of the global model. Unlike with local models, the user can anticipate the results of new predictions with perfect fidelity, i.e., the prediction of the model as grasped by the user completely matches that of the actual model.

\section{Performance of kNN Trees as a function of $k$}
\label{sec:knntree}

The kNN Trees are used as a baseline method that, similarly to our meta trees method, transfers knowledge from other users to augment the limited training set that is available for each user. The results we report in the paper are for the best $k$ and maximal tree depth (found to be three for all three datasets) as evaluated on the test set, thus we provide an upper bound on the method's performance for each benchmark.

\begin{figure}[t]%
\centering
\begin{tabular}{c@{~}c@{~}c}
MovieLens-100k     &   MovieLens-1M  &   Jester       \\
\includegraphics[width=0.3102105001\textwidth]{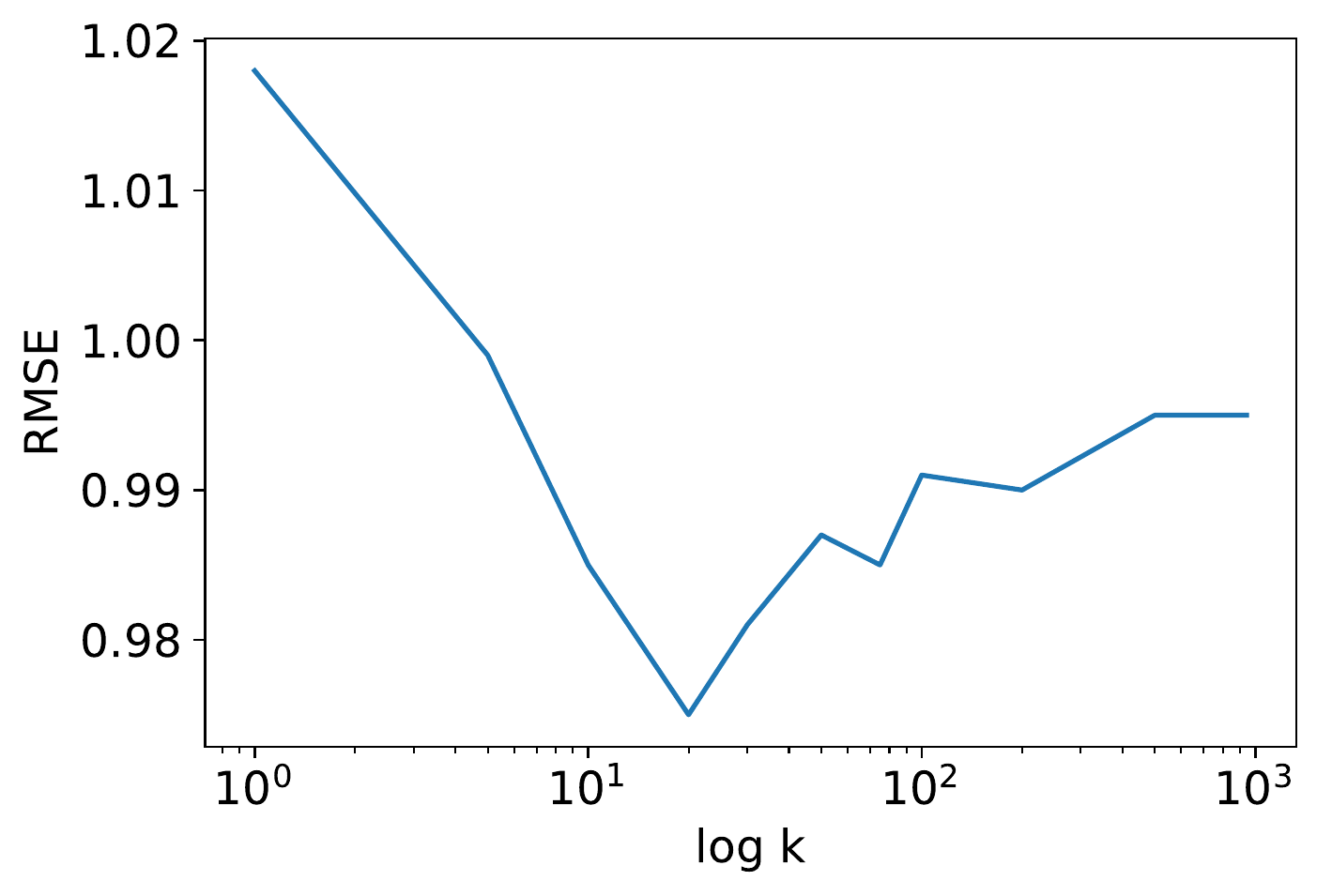}&
\includegraphics[width=0.3102105001\textwidth]{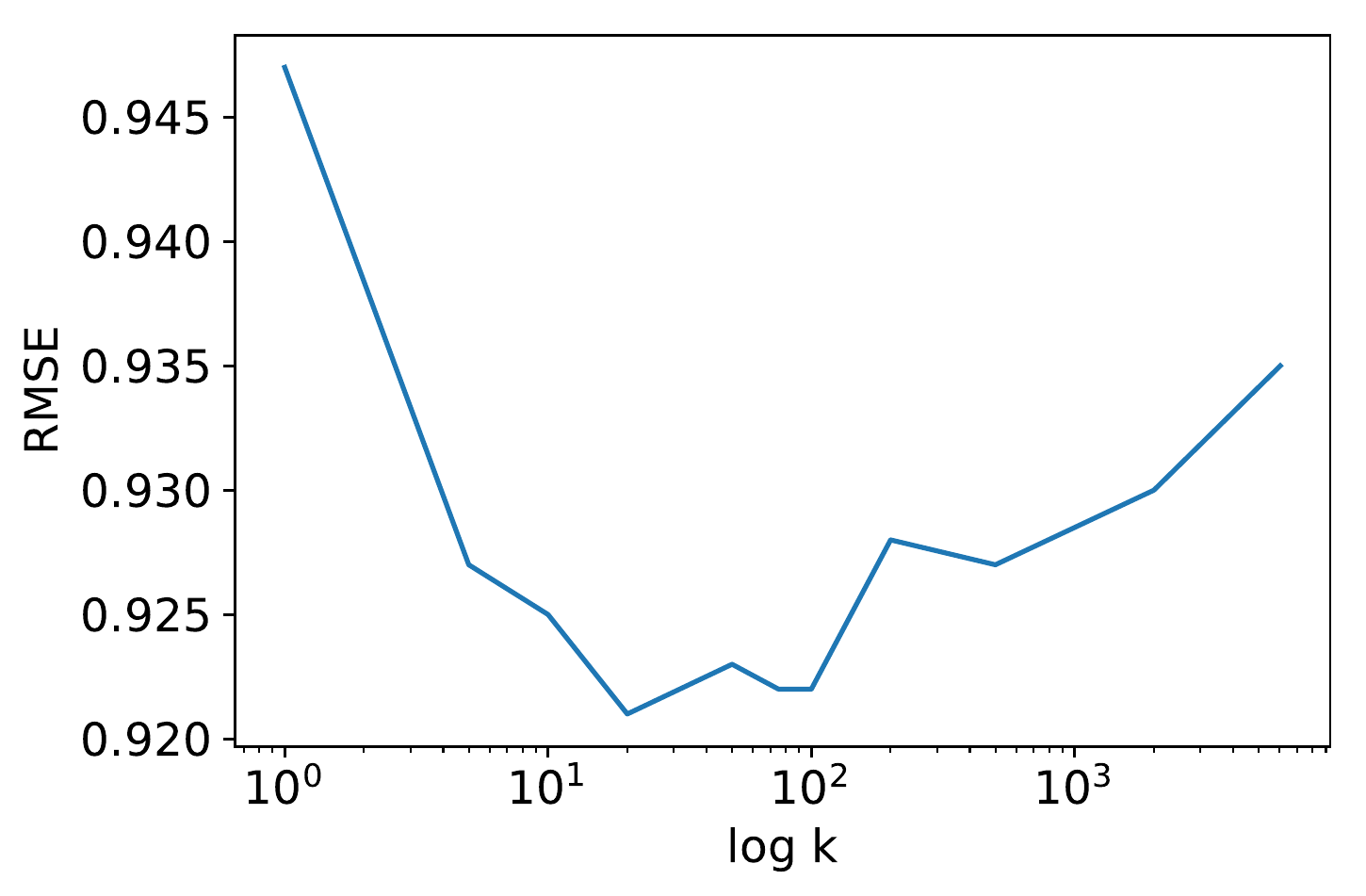}&
\includegraphics[width=0.3102105001\textwidth]{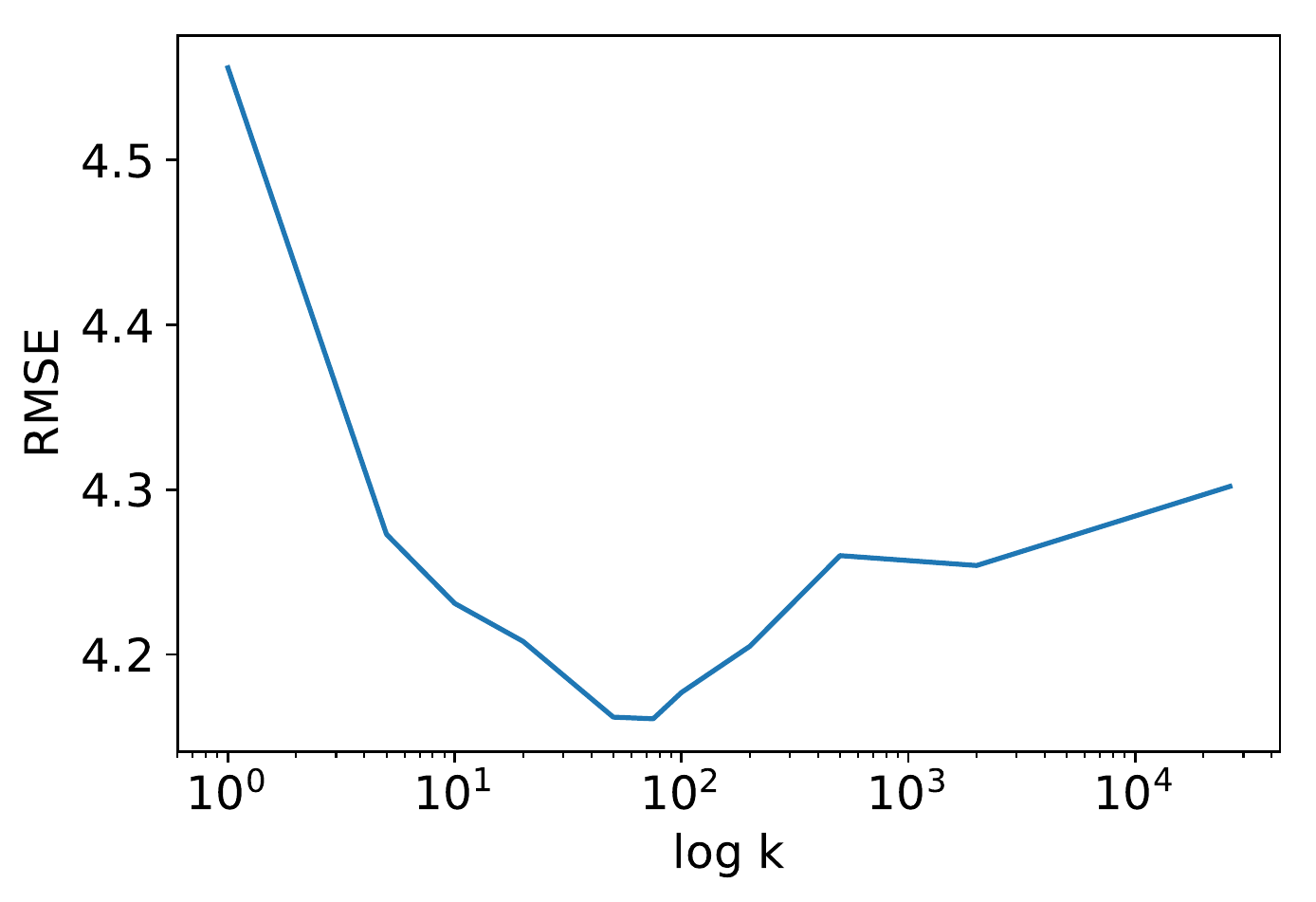}
\\
\end{tabular}
\caption{KNN Trees performance as a function of $k$ on all three datasets.}
\label{fig:knntrees}
\end{figure}

In order to provide a fuller characterization of the performance of this baseline method, we provide the performance of the kNN Trees algorithm as a function of $k$ on all three datasets. Fig.~\ref{fig:knntrees} shows that when $k=1$ (Local Trees) the algorithm indeed does not perform very well, the performance improves as the single user training set is augmented with training data of its similar users. At higher $k$ values, the performance starts to degrade until finally the algorithm uses all training data for all users (Global Tree).

\section{Parameter Sensitivity}
\label{sec:paramsensitivity}
\begin{figure}[t]%
\centering
\begin{tabular}{c@{~}c@{~}c}
MovieLens-100k     &   MovieLens-1M  &   Jester       \\
\includegraphics[width=0.3102105001\textwidth]{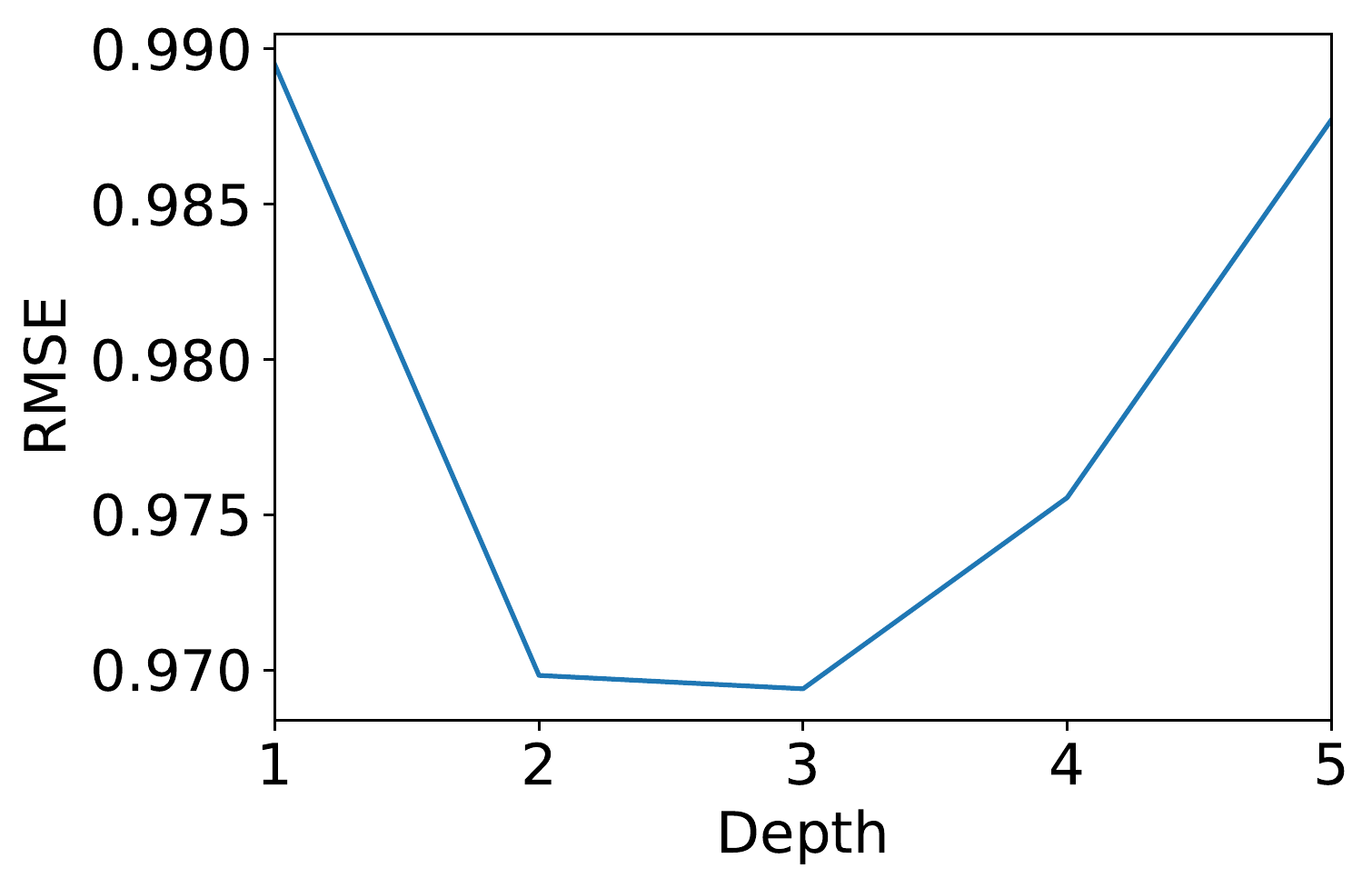}&
\includegraphics[width=0.3102105001\textwidth]{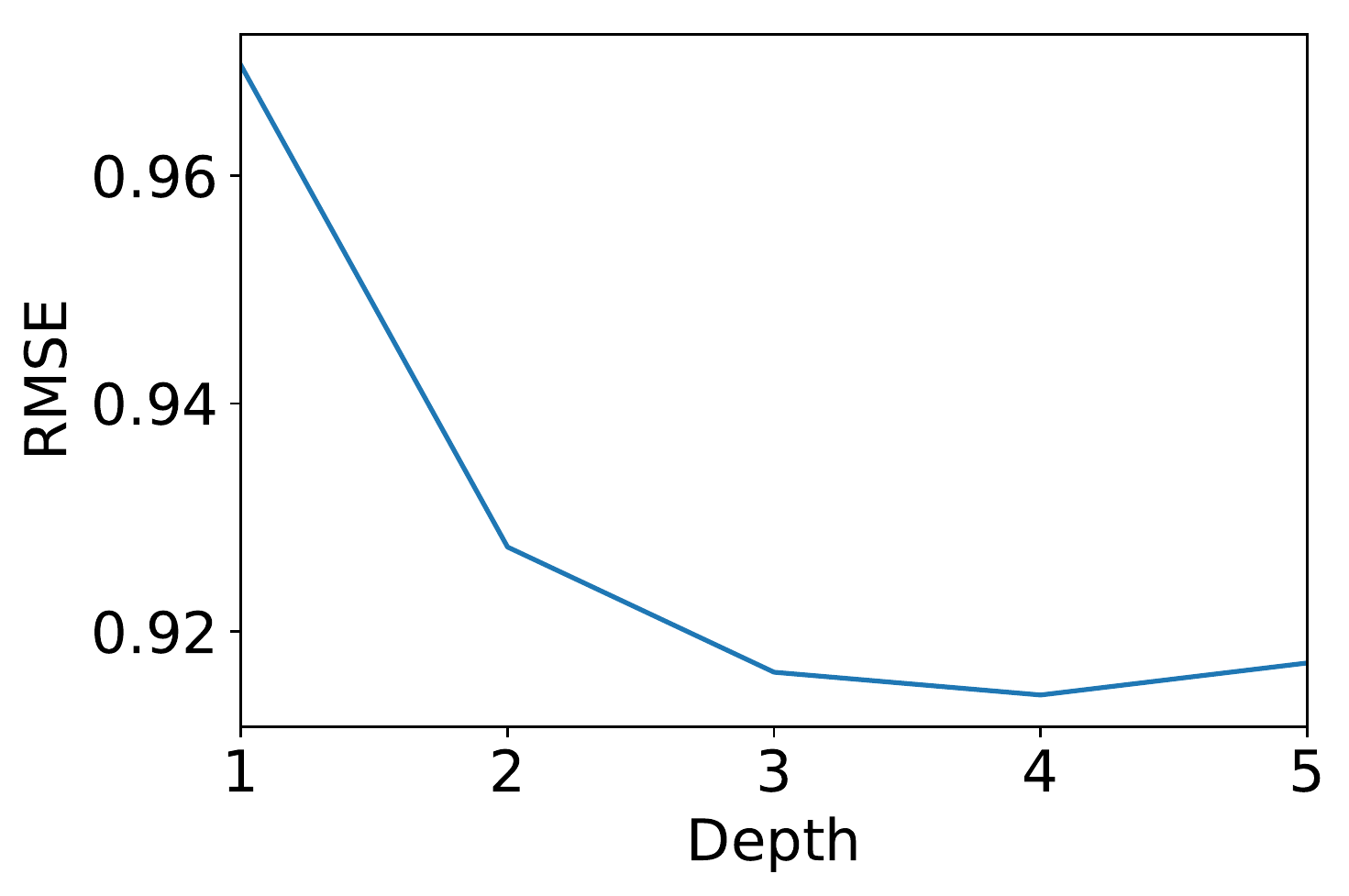}&
\includegraphics[width=0.3102105001\textwidth]{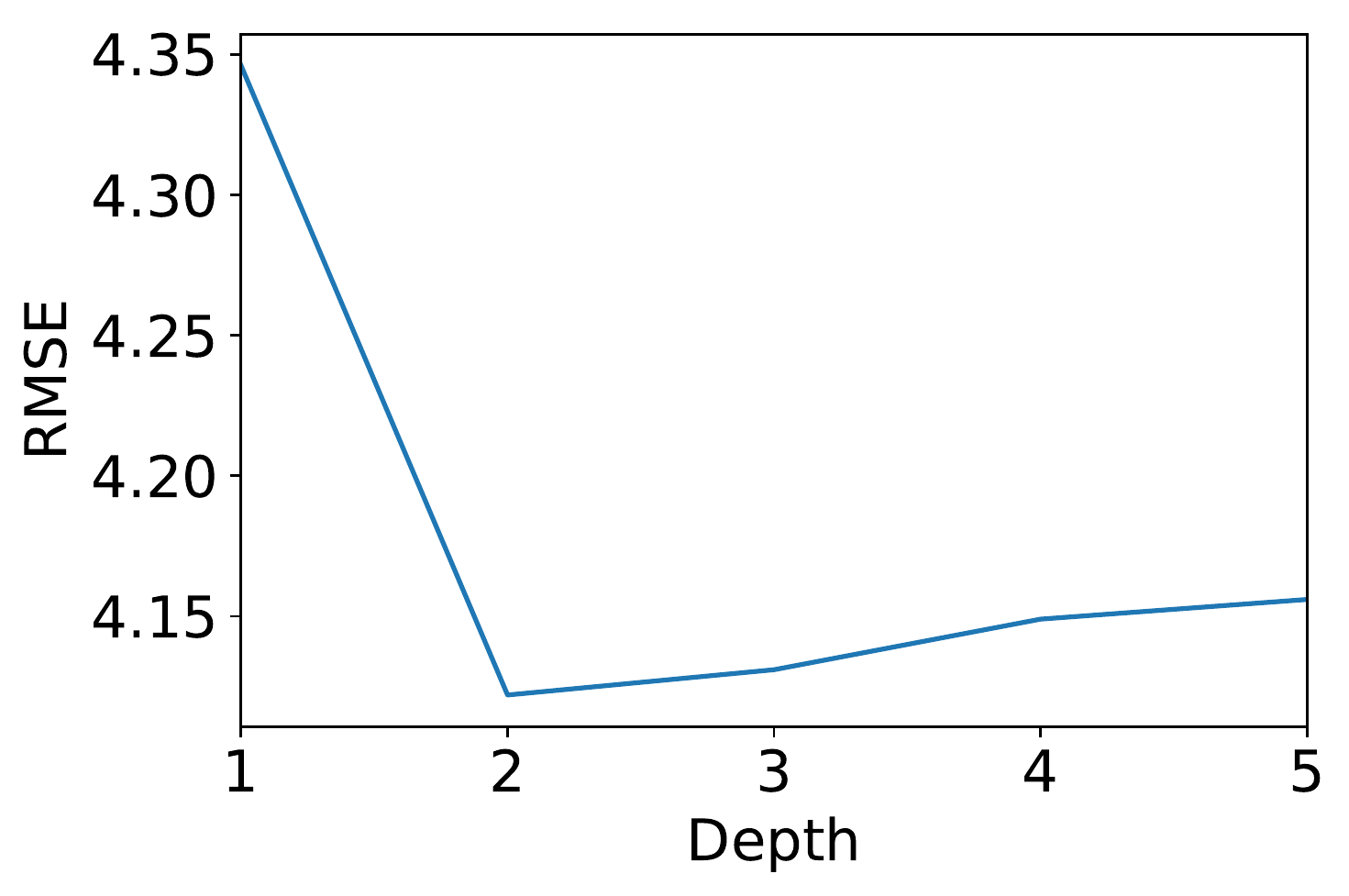}
\\
\includegraphics[width=0.3102105001\textwidth]{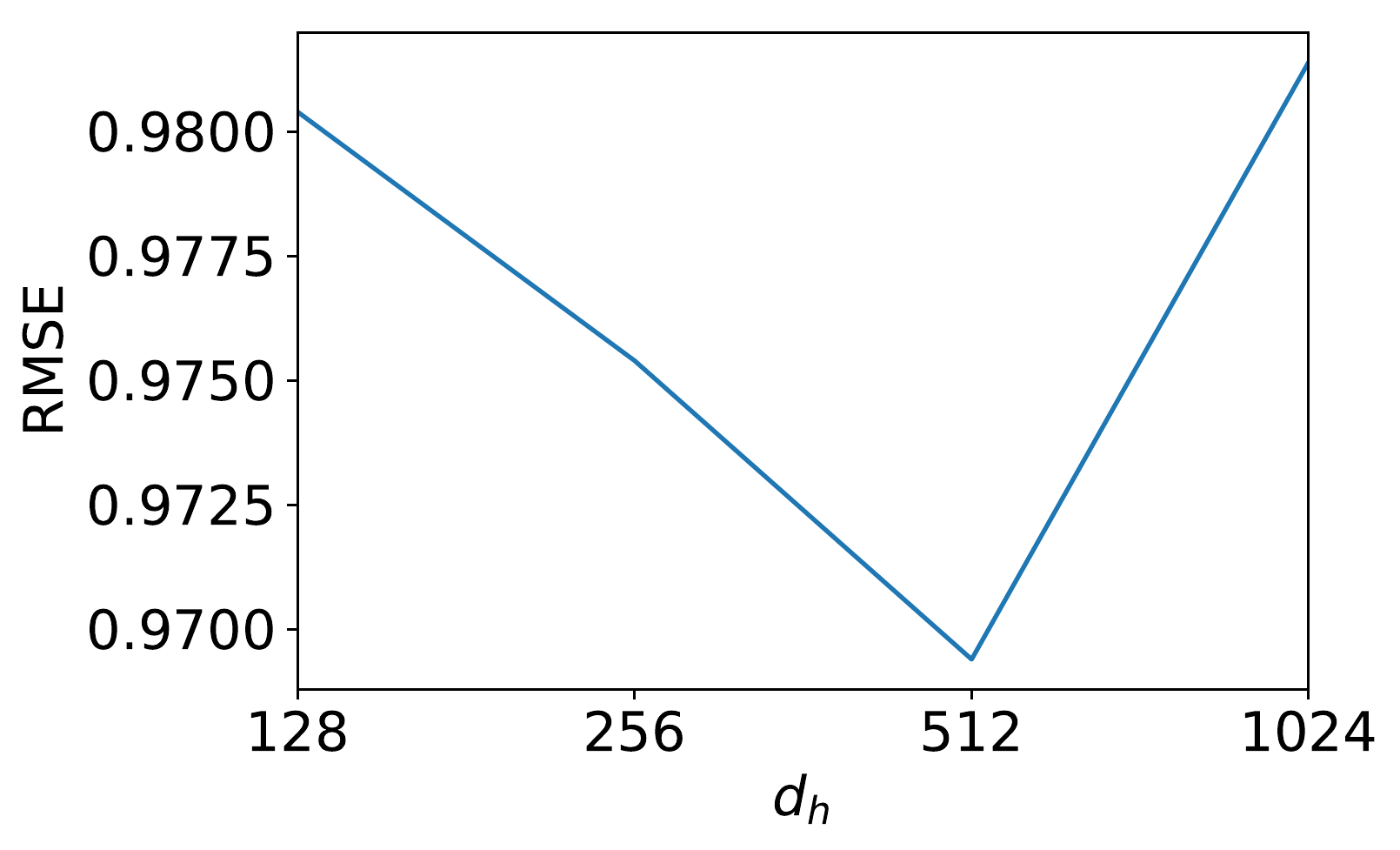}&
\includegraphics[width=0.3102105001\textwidth]{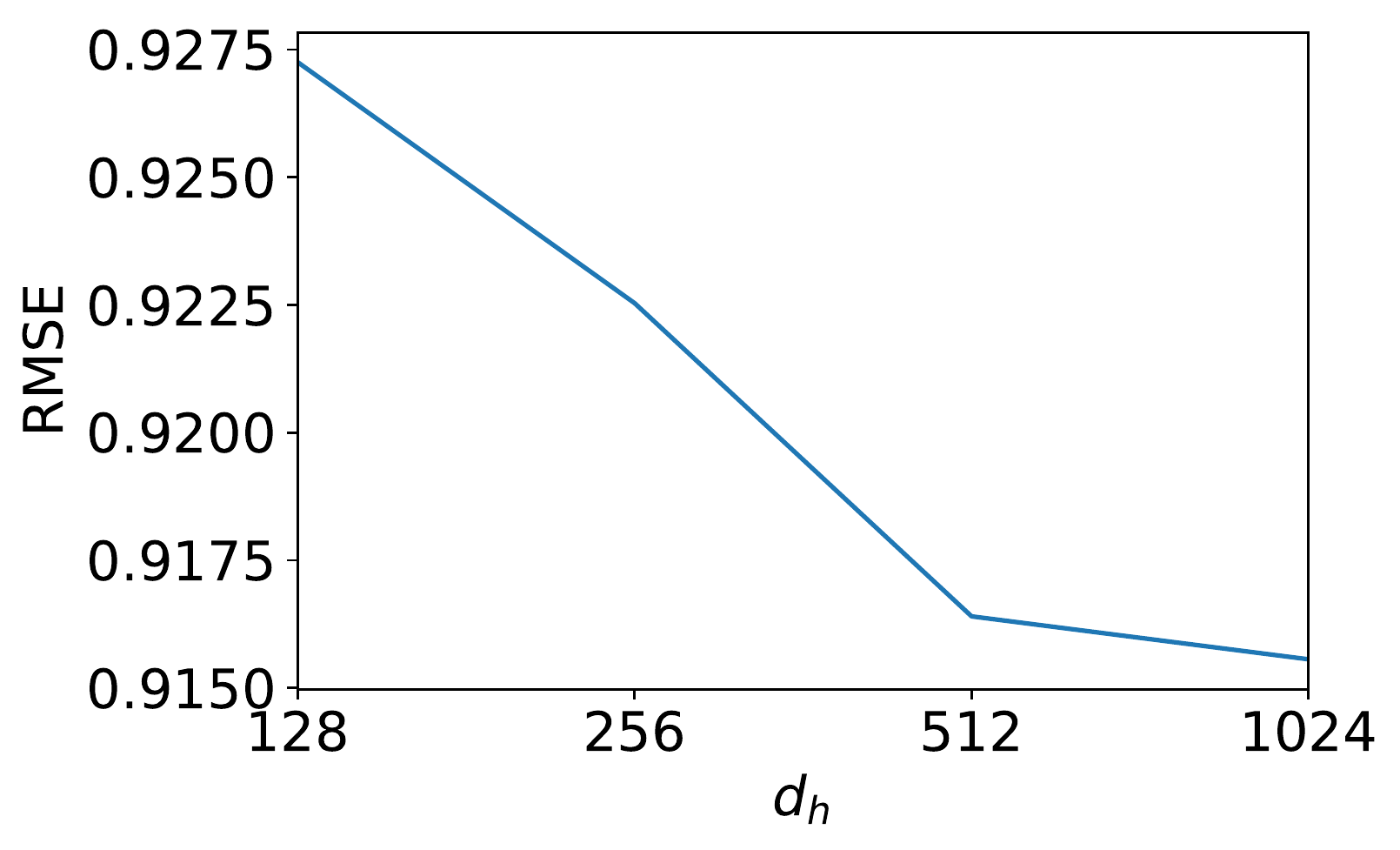}&
\includegraphics[width=0.3102105001\textwidth]{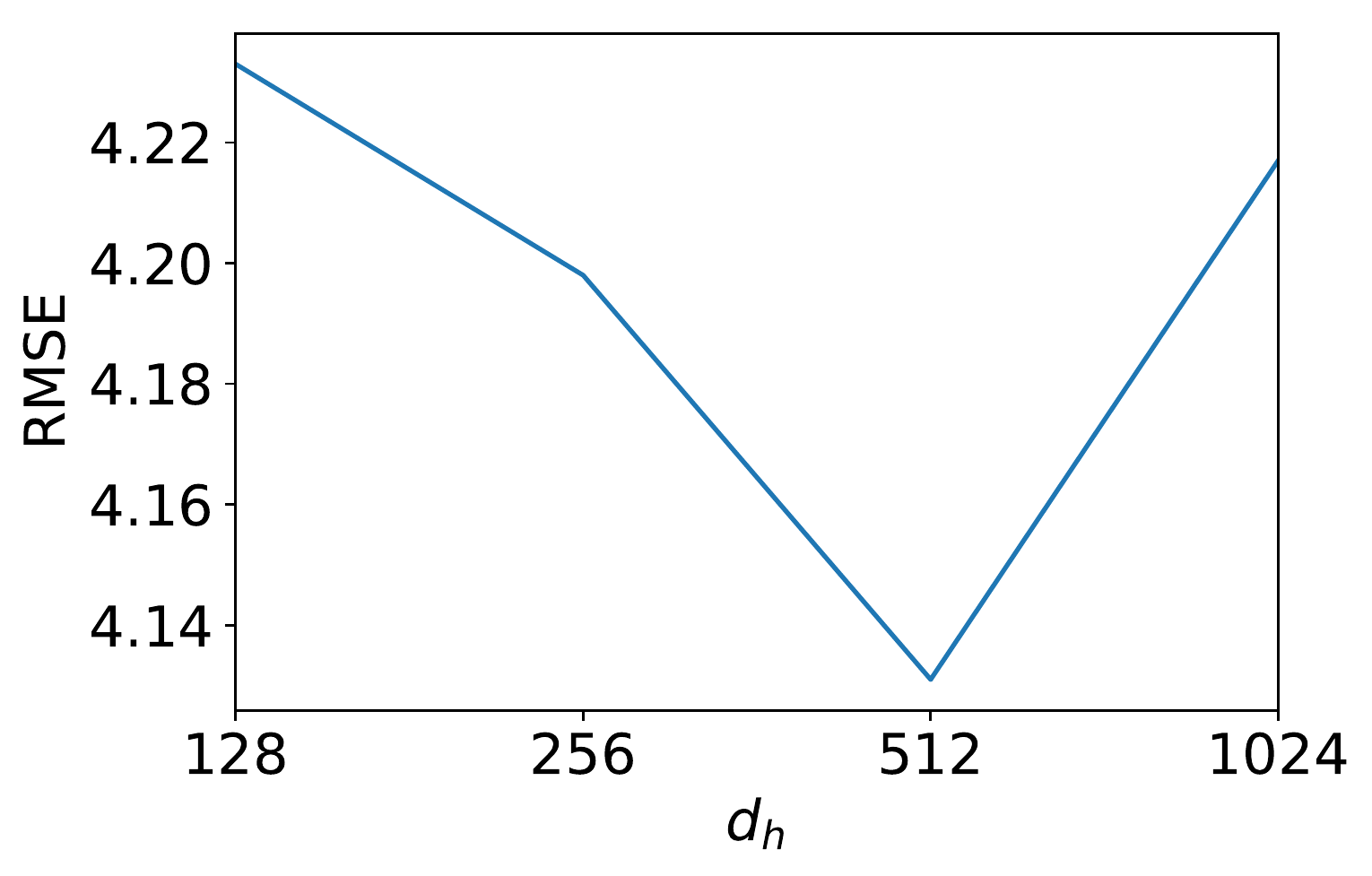}
\\
\end{tabular}
\caption{Model sensitivity to $d_h$ (top row) and tree depth (bottom) for the three datasets.}%
\label{fig:psense}%
\end{figure}

We test the sensitivity to the underlying parameters: maximal depth (for a fixed architecture) and the size of the embedding $d_h$. In our experiments, these were set early on to three and 512, respectively. Fig.~\ref{fig:psense} shows the effect on the RMSE score when varying either one of these. As can be seen, the method is largely insensitive to its parameters. 

The parameters presented in the main text are not optimal: gains in accuracy can be achieved, in some cases, by considering deeper or shallower trees or by enlarging the embedding.

\section{Robustness to Perturbations of the Training Set}
\label{sec:robust}

We can expect that a model that transfers knowledge between different users would be more robust to perturbations of the training set of a new user than a model that is learned from scratch. In addition, the mean pooling that we perform to compute the training set embedding $r = \frac{1}{n}\sum_{i=1}^n r_i$ averages all samples together, which also indicates robustness.

\begin{figure}[t]%
\centering
\begin{tabular}{c@{~}c@{~}c}
\includegraphics[width=0.3102105001\textwidth]{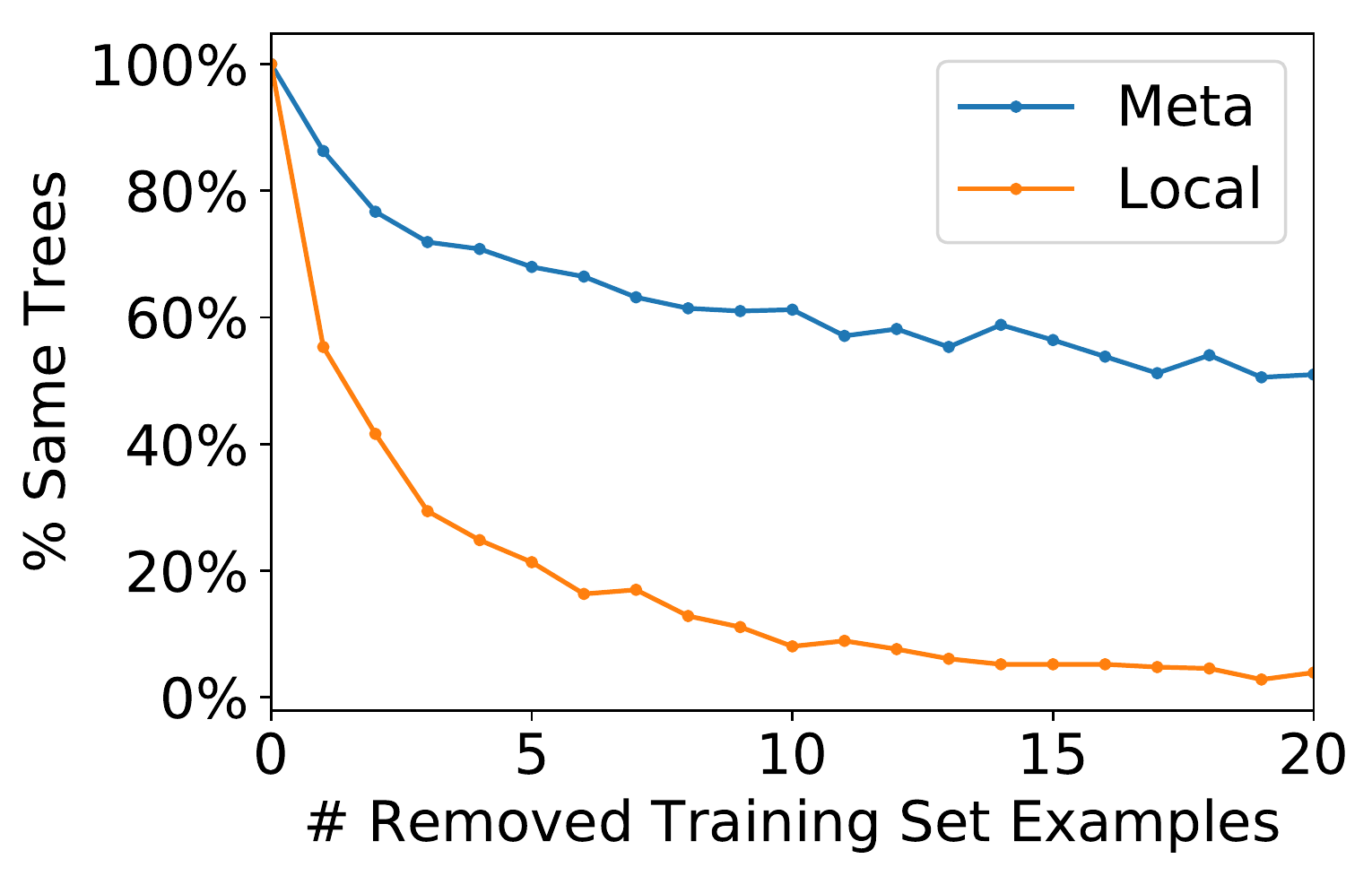}&
\includegraphics[width=0.3102105001\textwidth]{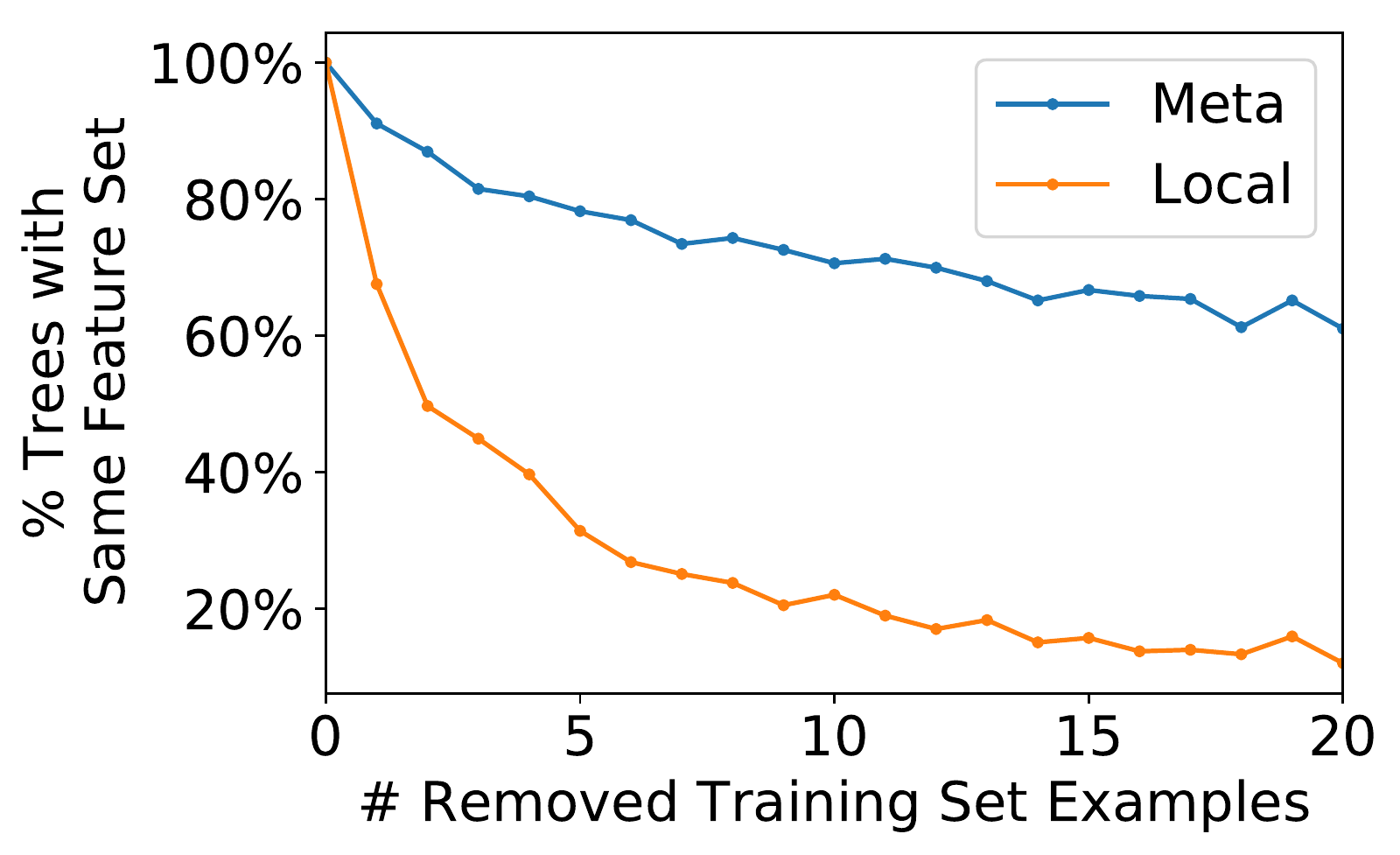}&
\includegraphics[width=0.3102105001\textwidth]{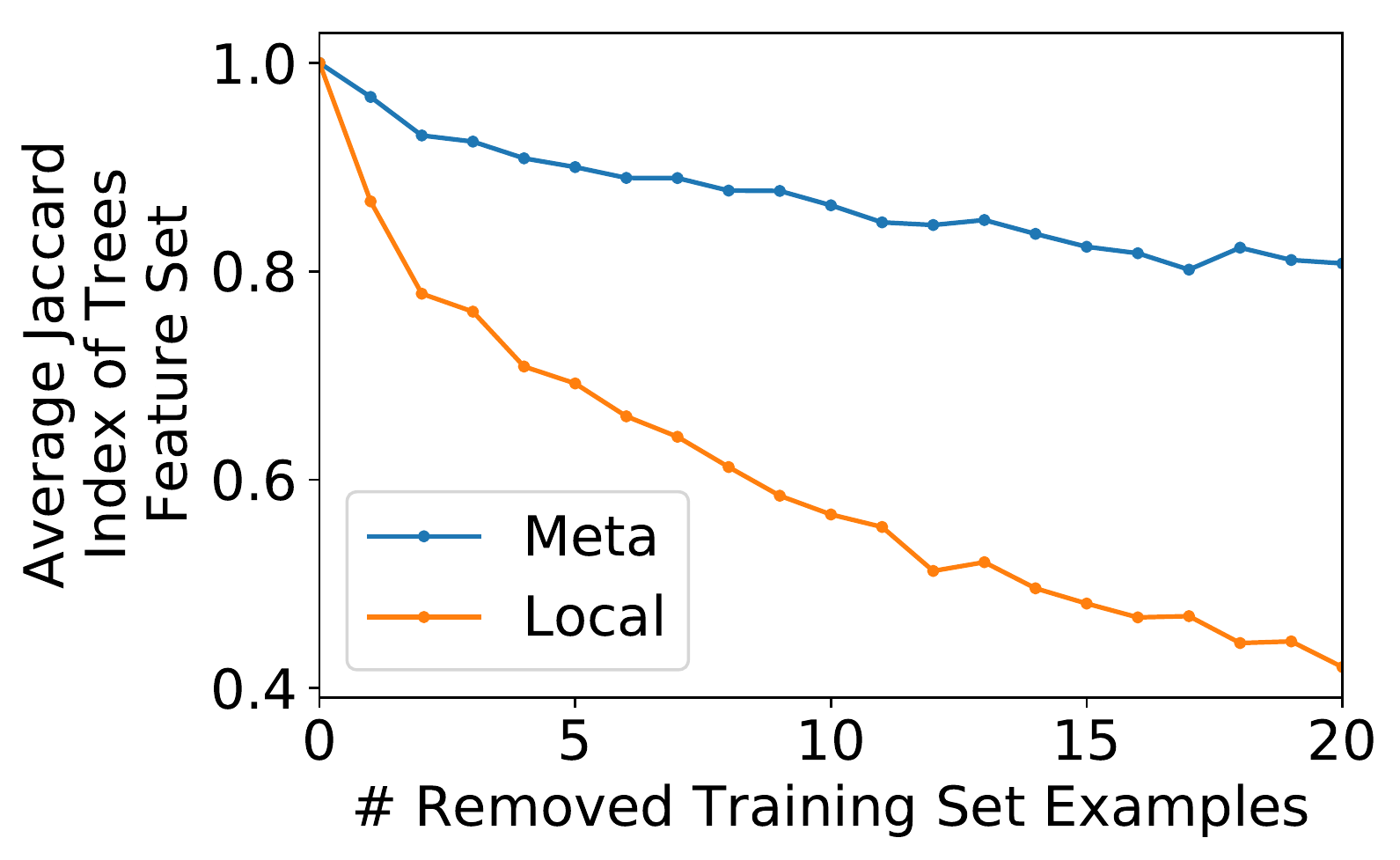}
\\
(a)&(b)&(c)\\
\end{tabular}
\caption{Measuring the robustness of trees generated by the meta-tree model (``Meta'') and the local-trees (``Local'') baseline. Comparing the trees generated using the original user's training set and after removing a random subset from it. (a) Exact match of the features used in each tree node (order must be identical). (b) Exact match of the set of features used. (c) The Jaccard index of the set of features used.}
\label{fig:robust}
\end{figure}

The robustness to perturbations of the training set is shown in Fig.~\ref{fig:robust}, in which we compare our meta tree model to the local model on the MovieLens 100K dataset. The plots show the behavior as we remove more and more samples from the training set, averaging over 10 random subsamples of the training set. 

We first measure if the obtained tree after removing samples is identical to the original tree in the features along the nodes and in the order in which they appear. We then perform a more liberal test, which compares the sets of features regardless of the node order and multiplicity of each feature. For the sets of features we report both the percent of cases in which the two sets (obtained from the full training set and the sampled one) are identical, as well as the Jaccard index.

As can be seen, the proposed meta-tree model is far more robust than the local tree model.

\section{Cold Start: Behavior for a Small User Training Set}
\label{sec:coldstart}
\begin{figure}[t]%
\centering
\begin{tabular}{c@{~}c}
\includegraphics[width=0.4902105001\textwidth]{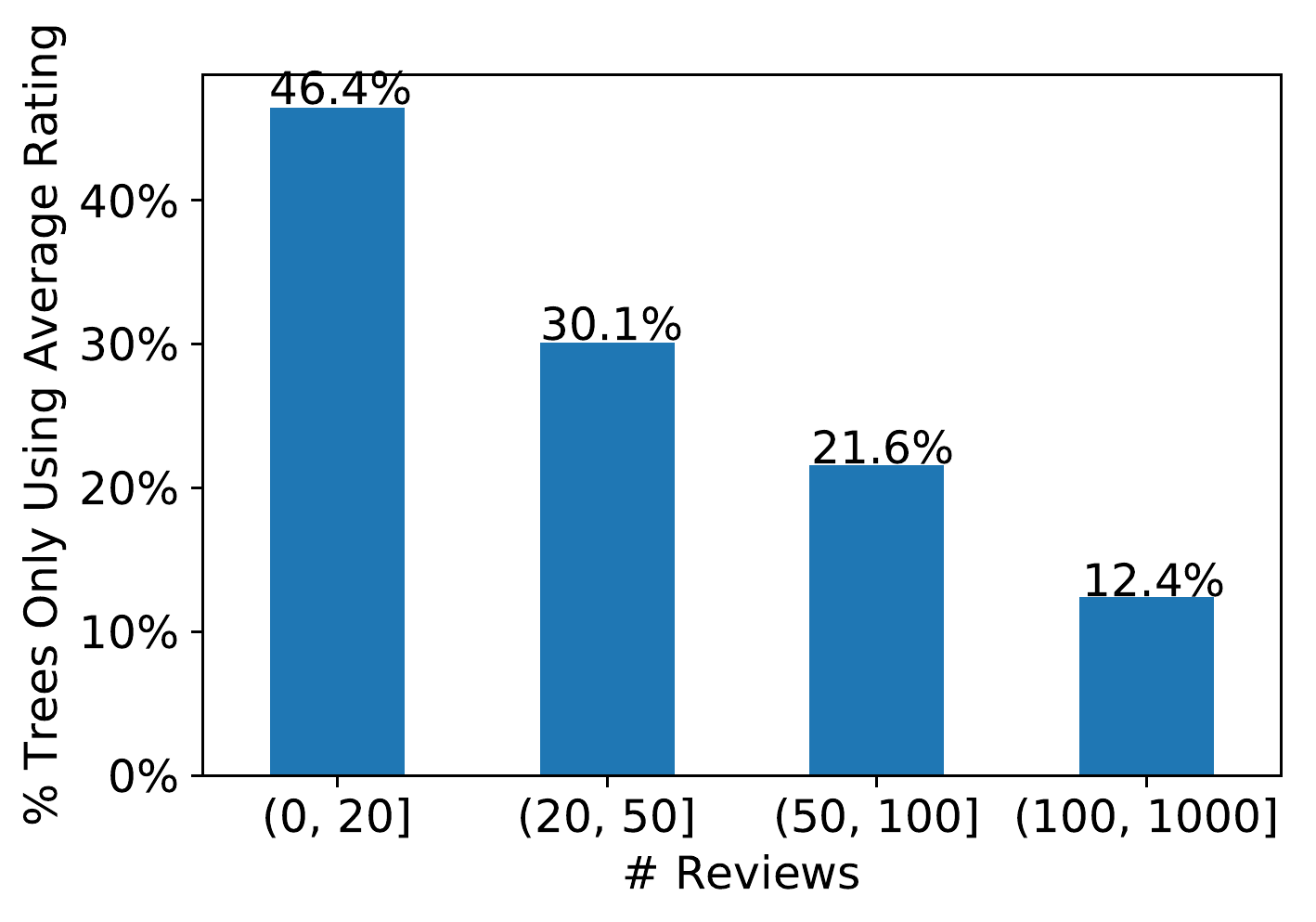}&
\includegraphics[width=0.4902105001\textwidth]{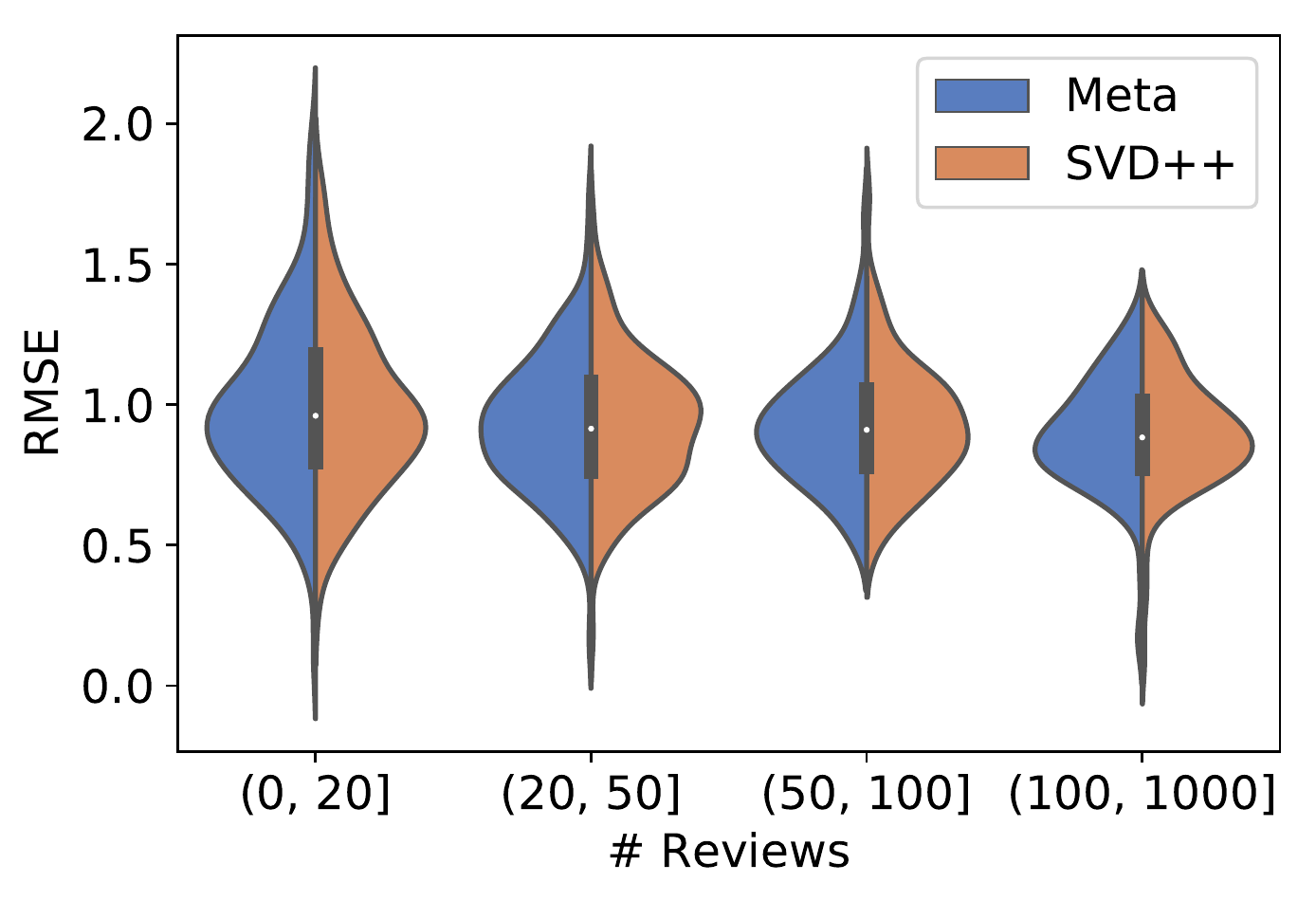}
\\
(a)&(b)\\
\end{tabular}
\caption{Analysis of the model behavior on the MovieLens-100k dataset. (a) Percentage of trees using only the item average rating as a function of the user's training set size. (b) Performance comparison of the meta-tree algorithm (``Meta'') and SVD++ for different per-user training set size.}%
\label{fig:coldstart}%
\end{figure}

As an example of the ability to gain insight from interpretable models, we study the model behavior with respect to the size of the user training set. In Fig.~\ref{fig:coldstart}(a) we observe that the algorithm tends to produce more trees using only the item average rating for users with small training sets. This observation makes sense as for such users it is harder to have high certainty regarding the users' preferences hence using the population consensus (with adjustments) is a reasonable option. From  Fig.~\ref{fig:coldstart}(b) we learn that the algorithm is performing similarly to SVD++ across all bins of the training set size for test users. 

\section{Odd Users: the Case of Contrarian Ratings}
\label{sec:contrarian}

\begin{figure}[t]%
\centering
\includegraphics[width=0.6\textwidth]{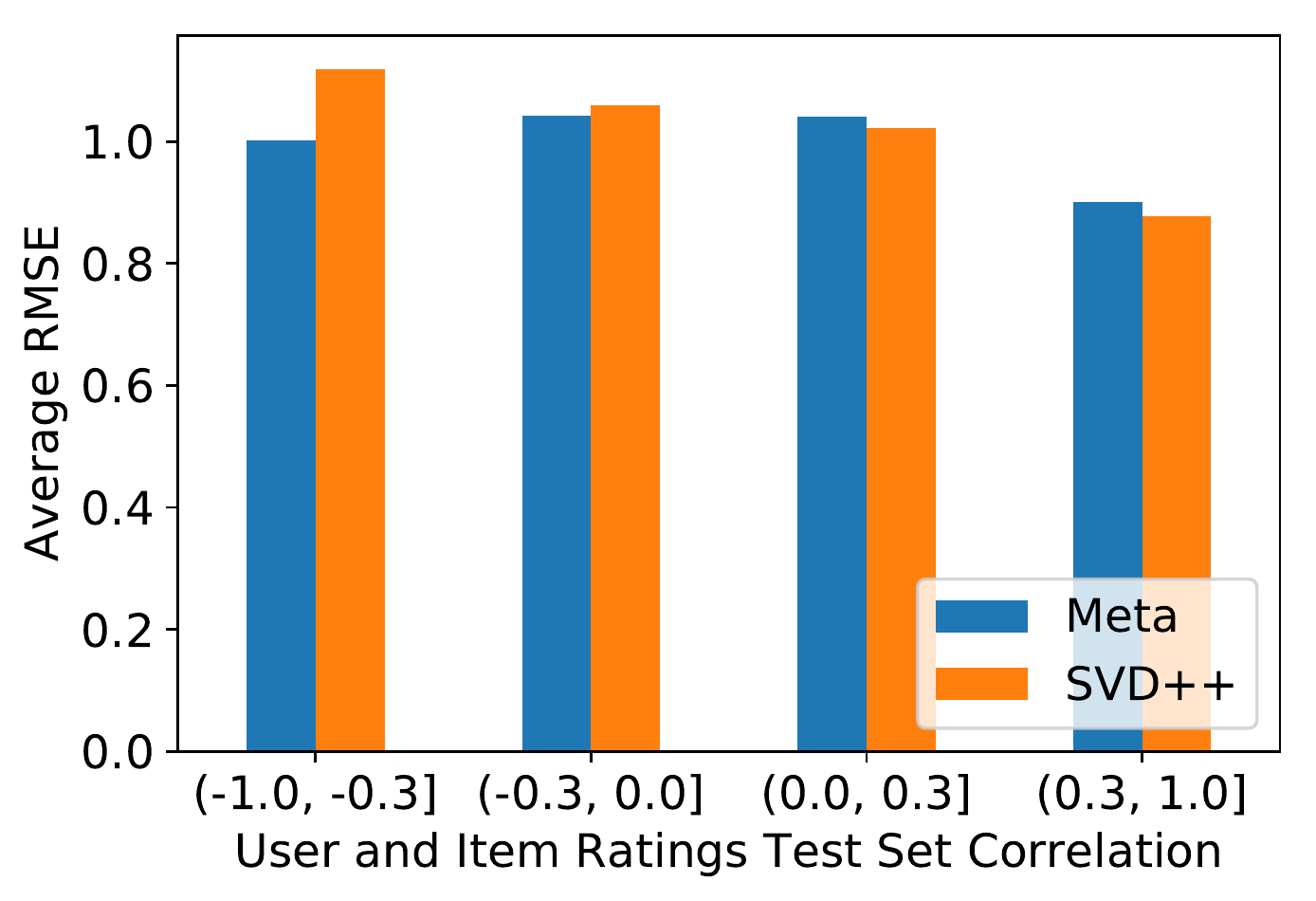}
\caption{Comparison of the meta-tree model (``Meta'') and the SVD++ algorithm performances on the MovieLens-100k dataset with respect to the users' ratings correlation with the average item rating.}%
\label{fig:contrarian}%
\end{figure}

Inspecting the trees our algorithm produces, we encountered an unexpected type of trees generated across all three datasets. We call these trees ``Reverse trees'', and their logic indicates that the users represented by them tend to prefer items which other users rate with low scores. This logic is in contrast to the item bias term in CF algorithms, which is used to reduce the item average ratings from each rating. In order to test whether the data supports this model behavior, we compared the model performance to that of the SVD++ algorithm with respect to the correlation between the user's ratings and average item ratings. Fig~\ref{fig:contrarian} shows that our model indeed outperforms SVD++ for users with negative correlation and that the performance of SVD++ improves as the correlation increases, this phenomena was, to our knowledge, unknown and is a demonstration of the advantages of having such an interpretable model at hand.

\end{document}